\documentclass[11pt]{article}

\usepackage{amssymb,amsmath}
\usepackage{float}
\usepackage{graphicx}
\usepackage{caption}
\usepackage{array}

\ifdefined\XeTeXversion\else
  \usepackage[utf8]{inputenc}
\fi
\usepackage[T1]{fontenc}
\usepackage{booktabs}
\usepackage{microtype}

\usepackage{xcolor}
\definecolor{hunyuanblue}{HTML}{1E4A8F}
\usepackage{pifont}

\usepackage{hunyuan}

\usepackage[capitalize,noabbrev]{cleveref}
\usepackage{tabularx}
\usepackage[skins]{tcolorbox}
\usepackage{xspace}
\usepackage{fancyhdr}
\usepackage{xurl}

\makeatletter
\long\def\@makecaption#1#2{%
  \vskip 10pt
  \setbox\@tempboxa\hbox{#1: #2}%
  \ifdim \wd\@tempboxa >\hsize
    \noindent #1: #2\par   %
  \else
    \hbox to\hsize{\hfil\box\@tempboxa\hfil}%
  \fi}
\makeatother

\hypersetup{
  pdftitle={Game-Creation-Agents: Autonomous Game Production},
  pdfauthor={Multimodal Reinforcement Learning Team, Tencent Hy},
  colorlinks=true,
  breaklinks=true,
  linkcolor=hunyuanblue,
  citecolor=hunyuanblue,
  urlcolor=hunyuanblue,
}

\makeatletter
\def\section{\@startsiction{section}{1}{\z@}{-0.24in}{0.10in}
             {\large\bf\raggedright\color{hunyuanblue}}}
\def\subsection{\@startsection{subsection}{2}{\z@}{-0.20in}{0.08in}
                {\normalsize\bf\raggedright\color{hunyuanblue}}}
\makeatother

\fancypagestyle{plain}{%
  \fancyhf{}%
  \fancyfoot[C]{\thepage}%
}

\fancypagestyle{firststyle}{%
  \fancyhf{}%
  \fancyfoot[C]{\thepage}%
}

\usepackage{longtable}
\usepackage{colortbl}
\usepackage{needspace}
\usepackage{placeins}
\usepackage{float}
\usepackage{flafter}
\usepackage{textcomp}
\usepackage{makecell}
\DeclareRobustCommand{\gcaid}[1]{%
  {\ttfamily\hyphenchar\font=-1
   \exhyphenpenalty=0 \relax
   \def\_{\textunderscore\allowbreak}%
   #1}}

\newcommand{\gca}{\mbox{GCA}\xspace}
\newcommand{\agentname}[1]{\textsf{#1}}
\newcommand{\toolname}[1]{\gcaid{#1}}
\newcommand{\gatename}[1]{\gcaid{#1}}
\newcommand{\cfgkey}[1]{\gcaid{#1}}
\newcommand{\artifact}[1]{\gcaid{#1}}


\definecolor{yesgreen}{RGB}{46,139,87}
\definecolor{oursband}{RGB}{232,240,247}
\newcommand{\cyes}{\textcolor{yesgreen}{\ding{51}}}

\newcommand{\ghlogo}{%
  \raisebox{-0.16em}{%
    \makebox[1.16em][c]{%
      \includegraphics[height=1.08em]{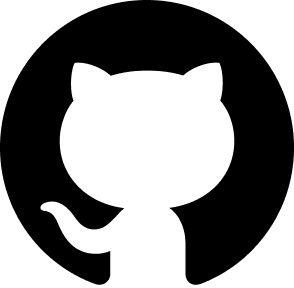}}}}
\newcommand{\reslink}[3]{\href{#2}{\mbox{#1\,{\small #3}}}}

\definecolor{abstractbg}{HTML}{F0F7FC}
\begin{document}

\thispagestyle{firststyle}
\vspace*{0.25cm}
{\color{hunyuanblue}\hrule height 0.6pt}
{\centering
{\vskip 6mm
\LARGE\bfseries Game-Creation-Agents: Autonomous Game Production\par}
\vskip 5mm
}
\vskip 2.5mm
{\color{hunyuanblue}\hrule height 0.6pt}
\vskip 3.5mm
{\centering
{\small\bfseries Multimodal Reinforcement Learning Team, Tencent\par}
}
{\centering
\reslink{\ghlogo}{https://github.com/Tencent/Game-Creation-Agents}{https://github.com/Tencent/Game-Creation-Agents}
\par}
\vskip 2mm

\begin{tcolorbox}[
  colframe=abstractbg,colback=abstractbg,boxrule=0pt,arc=2mm,enhanced,
  top=8pt,bottom=8pt,left=15pt,right=15pt,width=\textwidth,
  before skip=0pt,after skip=0pt]
\textbf{Abstract.}\quad
Game-Creation-Agents (\gca) turns a natural-language request into a
runnable Phaser~4 browser game. The pipeline first builds a multimodal game
design document that couples mechanics and scene contracts to grounded visual
references. It then projects that document into a runtime scaffold,
machine-readable contracts, and deterministic tests. Eight specialized agents
next carry out production: scenes are implemented serially, assets are produced
in parallel, and one reduced manifest joins the two branches into a single
build. Build-linked real-UI checks and a code-blind critic then verify that
build and decide whether the game may ship. Each blocker returns to the agent
that owns the affected artifact for a bounded repair, and the game is released
once all the gates pass. We evaluate finished games separately, through their rendered
interfaces alone, over prompt-matched tasks run under a common protocol. \gca{}
achieves stronger requirement satisfaction, core mechanics, content depth, and
functional visuals than a comparable open pipeline. In a side-by-side human
study, raters also score \gca{}'s games higher on every dimension and prefer
them overall.
\end{tcolorbox}

\noindent\begin{minipage}{\textwidth}
\setlength{\parskip}{0pt}
\captionsetup{font=small,hypcap=false}
\makeatletter
\long\def\@makecaption#1#2{%
  \vskip 5pt
  \setbox\@tempboxa\hbox{#1: #2}%
  \ifdim \wd\@tempboxa >\hsize
    \noindent #1: #2\par
  \else
    \hbox to\hsize{\hfil\box\@tempboxa\hfil}\par
  \fi}
\makeatother
\centering
\includegraphics[width=\linewidth]{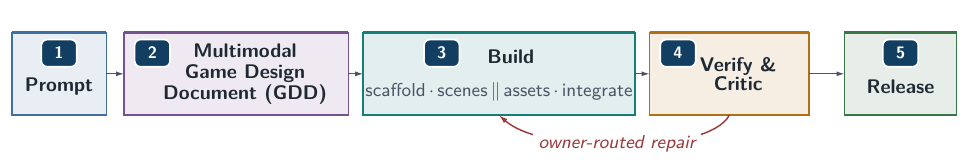}
\captionof{figure}{\gca{} pipeline. A prompt becomes a design document, then a
build that must pass verification before release; blockers route back to their
owners.}
\label{fig:pipeline}
\vspace{2pt}
\centering
\includegraphics[width=\linewidth]{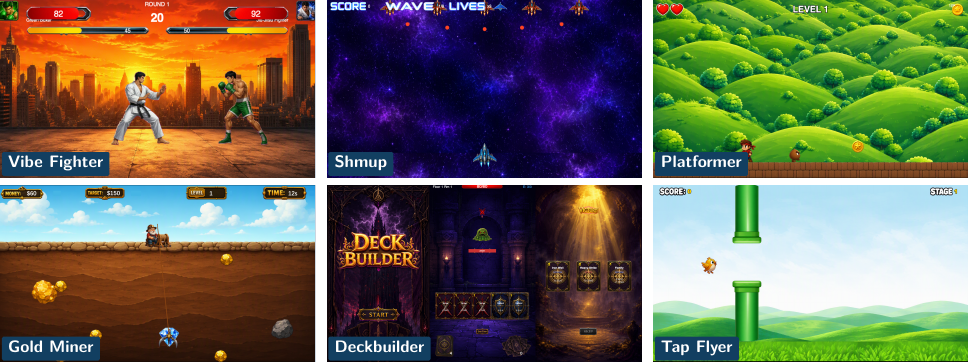}
\captionof{figure}{Six example games generated by \gca{}, spanning distinct
mechanics and visual styles.}
\label{fig:teaser-games}
\end{minipage}\par

\section{Introduction}
\label{sec:intro}

Large language model agents can now write and revise real programs, and recent
work organizes them into pipelines where each agent takes on a
role~\citep{qian2024chatdev,hong2024metagpt,wu2024autogen}. Games are a good
target for this, and a hard one. A game is not finished when it compiles and
renders. It has to take input, follow its own rules, move through levels and
end states, and show the player what is going on, using art that was generated
along with the code. One request therefore asks for dozens of things that can
be checked, spread across the design, the running code, and the assets.

The hard part is that nothing outside the pipeline tells it when it is done.
Code generation usually runs against a test suite that already
exists~\citep{chen2021codex,austin2021program,jimenez2024swebench}, but a game
pipeline has to build its own contracts, tests, and runtime evidence before it
can judge what it made. This is the test-oracle problem~\citep{barr2015oracle}.
An agent reporting that a scene works is not evidence that it
works~\citep{huang2024selfcorrect}, and a page that renders can still have
broken input, collisions, progression, end states, or interface feedback.
Running work in parallel makes it worse: design choices constrain scenes and
assets, code and art are produced at the same time but have to match at
integration, a repair can touch a file another agent owns, and leftovers from
an earlier attempt can look like current output. Without a record of which
artifact belongs to which agent and to which build, it is easy to mistake work
being done for work being finished.

\gca{} takes a game request in plain language and produces a runnable
Phaser~4 browser game~\citep{phaser2025}, along with its design package,
generated assets, executable tests, and verification evidence. Eight agents
split the work, each owning one kind of artifact. The design agent writes a
design specification paired with grounded scene concepts, detections, and
element crops; together these are the \emph{multimodal game design document
(GDD)}, or \emph{multimodal GDD}, and everything downstream is built from it.
Scenes are then implemented one at a time, while independent asset batches run
in parallel. \Cref{fig:pipeline} gives the whole sequence, and
\cref{fig:teaser-games} shows six games it produced.

Verification then runs on the assembled game, not on what the agents said
about it, and every check is bound to the build it ran against. Because each
artifact has exactly one writer, every blocker has a responsible owner, and
the repair goes back to that agent rather than to whoever noticed it. A typed
Model Context Protocol (MCP) control plane~\citep{anthropic2024mcp} enforces
both of these rather than leaving them to the agents.

We evaluate the finished games from outside the pipeline. The evaluator gets
only the prompt and one finished game through its rendered interface. It never
sees the multimodal GDD, the trace, the reports, or the critic's verdict, so
the internal gates cannot affect the score. Over prompt-matched tasks scored
under one protocol, \gca{} has higher mean Requirement Satisfaction, Core
Mechanics, Content Depth, and Functional Visuals than an open-source
baseline~\citep{jiang2026opengame}, and performs comparably on
Art \& Presentation. A human study of the
same pairs gives \gca{} higher mean ratings on all five dimensions and a clear
majority of decisive preferences.

The rest of the report is organized as follows. \Cref{sec:related} places \gca{} among prior
work on complete-game generation, evaluation and testing, and content
generation. \Cref{sec:system} details the pipeline and
\cref{sec:controlplane} the MCP control plane. \Cref{sec:evaluation} reports
the automated and human evaluations together with the Gold Miner case, and
\cref{sec:conclusion} summarizes the findings.

\section{Related Work}
\label{sec:related}

We position \gca{} by distinguishing the artifact each system produces and the
role assigned to execution evidence. \Cref{tab:related} compares
representative systems besides \gca{} through primary scope, evidence, and
outcomes. \gca{} introduces composite evidence gates that rise from a
seven-check scaffold boundary to the 39 checks that must all pass before the
final game ships.

\begin{table}[!ht]
\centering
\footnotesize
\setlength{\tabcolsep}{3.5pt}
\renewcommand{\arraystretch}{1.4}
\begin{tabularx}{\textwidth}{@{}
  >{\raggedright\arraybackslash}p{2.7cm}
  >{\raggedright\arraybackslash}p{3.0cm}
  >{\raggedright\arraybackslash}X
  >{\raggedright\arraybackslash}p{3.25cm}@{}}
\toprule
\textbf{System} & \textbf{Primary scope} &
\textbf{Executed or rendered evidence} & \textbf{Role of the outcome} \\
\midrule
GameGPT \citep{chen2023gamegpt}
  & Multi-stage game development
  & Task execution plus role-specific review
  & Workflow review and completion \\

OpenGame \citep{jiang2026opengame}
  & End-to-end Phaser game and asset creation
  & Build/test verification; companion benchmark executes the game
  & Iterative debugging; separately reported benchmark scores \\

MarioGPT \citep{sudhakaran2023mariogpt}
  & Text-conditioned Mario-level generation
  & Generated-level sampling and novelty search
  & Content generation and diversity \\
\addlinespace[2pt]

GameDevBench \citep{chi2026gamedevbench}
  & Godot development tasks
  & Engine validation; screenshots and runtime video studied as agent feedback
  & Development-agent benchmark \\

OpenGame-Bench \citep{jiang2026opengame}
  & Finished web games
  & Headless browser execution and multimodal judging
  & Build Health, Visual Usability, and Intent Alignment scores \\

GameCraft-Bench \citep{luo2026gamecraft}
  & Complete Godot games and replay traces
  & Replayed inputs, gameplay video and frames, hidden rubric
  & Core Mechanics, Content Depth, Functional Visuals, and Art scores \\

GBQA \citep{jiang2026gbqa}
  & Bug discovery in generated games
  & Interactive exploration and structured bug reports
  & QA-agent benchmark against verified bugs \\
\addlinespace[2pt]

Wuji \citep{zheng2019wuji}
  & Online combat-game testing
  & State exploration, coverage, and explicit oracles
  & Bug detection in an existing game \\

TITAN \citep{wang2025titan}
  & Commercial MMORPG testing
  & State abstraction, action traces, reflection, and LLM oracles
  & Diagnostic QA reports \\

Voyager \citep{wang2023voyager}
  & Minecraft play and skill acquisition
  & Environment feedback and executable skills
  & Open-ended agent learning \\
\midrule
\rowcolor{oursband}
\gca{} (ours)
  & End-to-end Phaser game and asset creation
  & Static, browser, play, layout, visual, and performance evidence
  & In-pipeline release gates; separate external evaluation \\
\bottomrule
\end{tabularx}
\caption{Representative game systems and the role of evidence.}
\label{tab:related}
\end{table}

\paragraph{Complete-Game Generation.}
GameGPT~\citep{chen2023gamegpt} is an early precedent for game-specific role
decomposition, organizing development into planning, task classification, code
generation, execution, and summarization with reviewer roles.
OpenGame~\citep{jiang2026opengame} is the closest artifact-level comparison to
\gca{}: it also targets complete Phaser web games from natural-language
specifications, separates design from implementation, treats asset synthesis as
part of creation, and includes a verification stage. The two differ in where
responsibility lives. OpenGame concentrates reusable project and repair
knowledge in accumulated skills and a game-specialized code model, whereas
\gca{} partitions files across eight agents and routes failed production checks
back to the owning role, keeping its browser, gameplay, layout, visual, and
performance evidence inside the production workflow rather than reporting it as
a score. Architectural distinctions are not quality evidence on their own, so
\cref{sec:evaluation} compares the finished artifacts under one external
protocol.

\paragraph{Game Evaluation and Testing.}
Execution-based benchmarks score what an agent builds by running it.
GameDevBench~\citep{chi2026gamedevbench} sets coding agents substantial Godot
tasks~\citep{godot2026} and reports lower success as visual complexity
increases, and OpenGame-Bench~\citep{jiang2026opengame} executes finished web
games in a headless browser, judging build health, visible usability, and
prompt alignment apart from one another.
GameCraft-Bench~\citep{luo2026gamecraft} is the protocol we borrow from most
directly: we adopt its four dimensions and its requirement for
interaction-conditioned evidence, adapt execution to browser games, and report
the dimensions separately. A parallel line tests games rather
than building them, from planted-bug exploration~\citep{jiang2026gbqa} to
oracles derived from commercial defects~\citep{zheng2019wuji} and long-horizon
diagnostic agents for commercial titles~\citep{wang2025titan}, while agents
acting through rendered interfaces show how far such exploration can
reach~\citep{zhang2025videogamebench,he2024webvoyager,zheng2024seeact,%
cheng2024seeclick,zhou2024webarena}. Voyager explores on a far larger budget,
though it acts through a scripting API, accumulating executable skills from
textual environment feedback rather than from a rendered
screen~\citep{wang2023voyager}. All of them work on a game that already exists,
with exploration budgets to match. \gca{} keeps the same separation between
exploration and oracles on a much smaller budget, running bounded
checks on a game it has just assembled in order to decide whether the pipeline
may advance.

\paragraph{Content and Asset Generation.}
Procedural content generation builds levels and other in-game content from
formal or learned models~\citep{shaker2016pcg,summerville2018pcgml}, and
language models extend this to text-conditioned tile-level generation with
novelty search for variation~\citep{sudhakaran2023mariogpt}. Such systems focus
on content within a game, whereas complete-game agents must also integrate
controls, state transitions, assets, build configuration, and presentation.
Latent-diffusion and image-to-video models widen the range of assets a creation
pipeline can synthesize~\citep{rombach2022ldm,blattmann2023svd}, but generation
alone does not show that an asset is usable in context, and multimodal judges
agree only imperfectly with human ratings~\citep{ku2024viescore}. We therefore
pair narrow asset and integration rubrics with deterministic size, reference,
layout, and runtime checks wherever those exist, and treat visual-model output
as one source of evidence rather than as ground truth.

\section{System Architecture}
\label{sec:system}

\gca{} is an artifact-producing pipeline. Each stage accepts named inputs from
its predecessors, writes a class of outputs that it alone owns, and records the
evidence its successor requires. A run starts from startup, design, optional
authored level design, and scaffold; implementation and asset production then
execute concurrently and join before asset integration; and verification
alternates with bounded, owner-routed repair until a final full pass clears
release. \Cref{fig:pipeline} gives this flow, and the rest of the section walks
it stage by stage, stating what each stage leaves behind.
\Cref{app:interfaces,app:gates} retain the exact dispatch schemas, gate
membership, and thresholds.

One rule governs every transition: a successful agent return does not advance
the run~\citep{huang2024selfcorrect}. The orchestrator validates that the
promised outputs exist, satisfy their contracts, and belong to the active
workspace, pipeline run, and build digest, and only then starts the next stage.
It controls dispatch, joins, state transitions, and gates while authoring no
game content itself; the substantive work belongs to the eight agents of
\cref{tab:agents}, each owning one artifact class. That split is what makes the
artifact graph explicit: the multimodal GDD defines intent, scaffold projects
contracts, implementation and asset agents realize disjoint outputs,
integration combines them, and verification records evidence about the
resulting runtime.

\begin{table}[!ht]
\centering
\small
\setlength{\tabcolsep}{5pt}
\renewcommand{\arraystretch}{1.40}
\begin{tabularx}{\textwidth}{@{}l>{\raggedright\arraybackslash}X>{\raggedright\arraybackslash}p{4.9cm}@{}}
\toprule
\textbf{Agent} & \textbf{Responsibility} & \textbf{Owned artifact class} \\
\midrule
\agentname{design} & Author the multimodal GDD: the design specification plus linked
concept mockups per screen and grounded crops of each element
& Design and concept evidence \\

\agentname{level-design} & Materialize or revise authored levels; dispatched
only when routing calls for them
& Authored level set \\

\agentname{scaffold} & Project the multimodal GDD into a Phaser~4 shell plus
the machine-readable contracts and deterministic tests that later gates read
& Runtime shell, contracts, and tests \\

\agentname{implement} & Implement one scene's gameplay and its feature tests;
dispatched once per scene, serially
& Scene implementations and feature tests \\

\agentname{assets} & Generate and register visual assets for one batch through
the control plane, with per-asset metadata sidecars
& Visual assets and metadata \\

\agentname{asset-integration} & Wire the finalized manifest into runtime
preloads, animations, and display parameters, then seal the integration receipt
& Runtime asset wiring and sealed receipt \\

\agentname{verify} & Run the verification DAG and fan its producers into one
identity-bound evidence report
& Verification evidence \\

\agentname{critic} & Judge shippability from evidence alone, having never seen
the source code
& Code-blind release verdict \\
\bottomrule
\end{tabularx}
\caption{Specialized agents and their owned artifacts.}
\label{tab:agents}
\end{table}

\paragraph{Startup.}
The orchestrator establishes the run's invariants without writing game content.
It preserves the request verbatim, selects one game directory, acquires an
exclusive workspace lease, derives immutable genre, interaction, quality, and
level routing, and persists the run identity. \emph{Produces}: the run
identity, the workspace lease, and a routing decision later stages cannot
revise. The stage validator therefore rejects files from a previous attempt,
and no later image or partial output can change the initial route.

\begin{figure}[!ht]
\centering
\includegraphics[width=\linewidth]{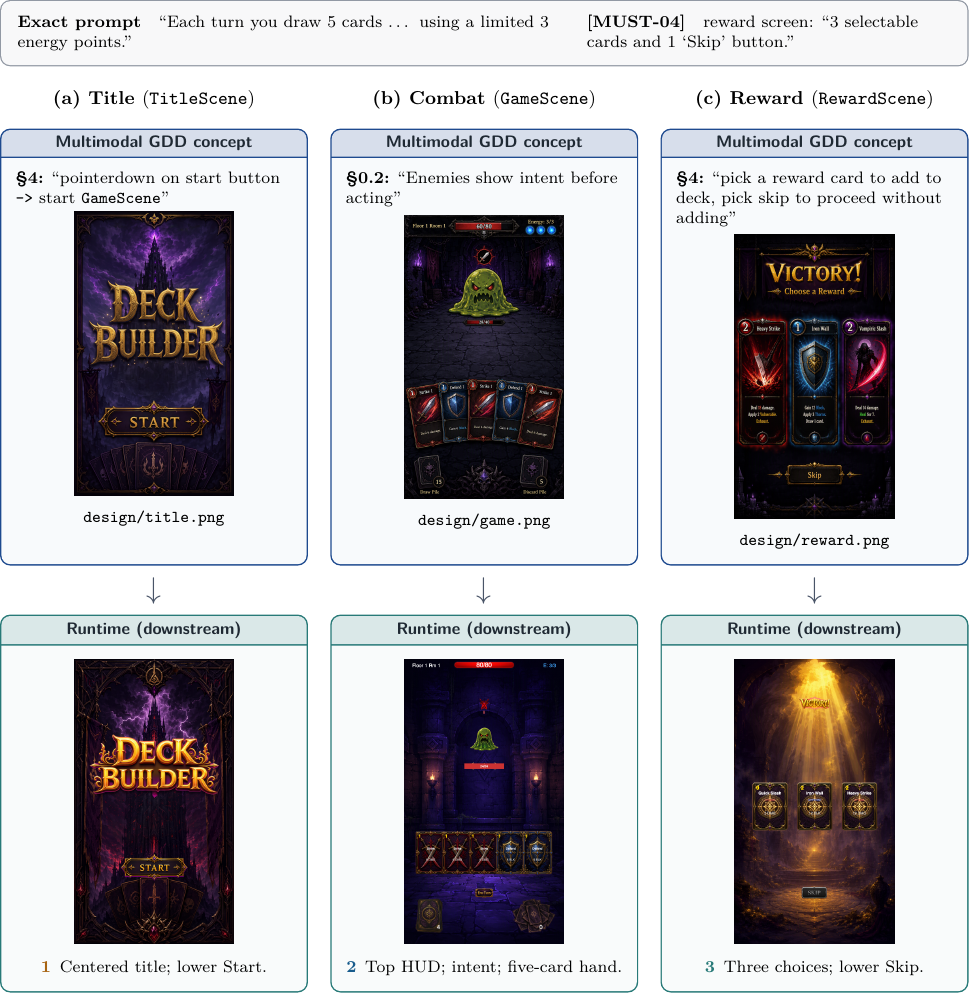}
\caption{Three Deckbuilder scene contracts link multimodal GDD
concepts to matching downstream runtime captures.}
\label{fig:design-package}
\end{figure}

\paragraph{Design.}
The design agent is the sole author of the design specification and its linked
visual evidence. Mechanics, state transitions, controls, progression, scene and
interface requirements, asset obligations, and testable objectives are paired
with screen concepts, normalized layout detections, and grounded element crops.
\emph{Produces}: the multimodal GDD, and the ordered scene list the
orchestrator extracts from it. The orchestrator validates that the document is
complete and that interactive scenes carry concept evidence or an explicit
bounded-failure waiver, but it does not repair the design itself.
\Cref{fig:design-package} illustrates the relation for the Title, Combat, and
Reward scenes of the game \gca{} produced for the Deckbuilder task of
\cref{tab:eval-corpus}.

The aligned runtime captures in \cref{fig:design-package} are downstream
outputs with the corresponding initial multimodal GDD. Their visual correspondence traces a scene
contract to its implementation, while it says nothing about the game rules like energy, turn
order, reward mutation, or deck conservation; those are behavioral claims that
only runtime interaction and state checks can settle. Keeping the two apart rejects a plausible visual design with buggy underlying implementation.

\paragraph{Level Design (Conditional).}
If routing selected an authored level model, the level-design agent writes and
validates the ordered stage files that progression requires. \emph{Produces}:
the level inventory, passed unchanged to scaffold and scene implementation.
Procedural, board, and other level-free games skip this stage, and incidental files
cannot alter routing.

\paragraph{Scaffold.}
Scaffold projects the accepted multimodal GDD into runnable structure.
\emph{Produces}: a Phaser~4 shell, scene stubs, an asset manifest, and
machine-readable interaction, objective, layout, rubric, and deterministic test
contracts. Because a different role writes them, these contracts are an oracle
for code that does not yet exist. A scaffold gate validates the outputs before
the server and build branches begin, and if a later design repair changes the
authority, scaffold reprojects only the stale paths named in its sealed
receipt.

\paragraph{Scene Implementation.}
Implementation runs serially over the declared scene list, because the scenes
share a code surface and serialization avoids hidden conflicts between them.
\emph{Produces}: scene code and the tests that exercise it.

\begin{figure}[!ht]
\centering
\includegraphics[width=\linewidth]{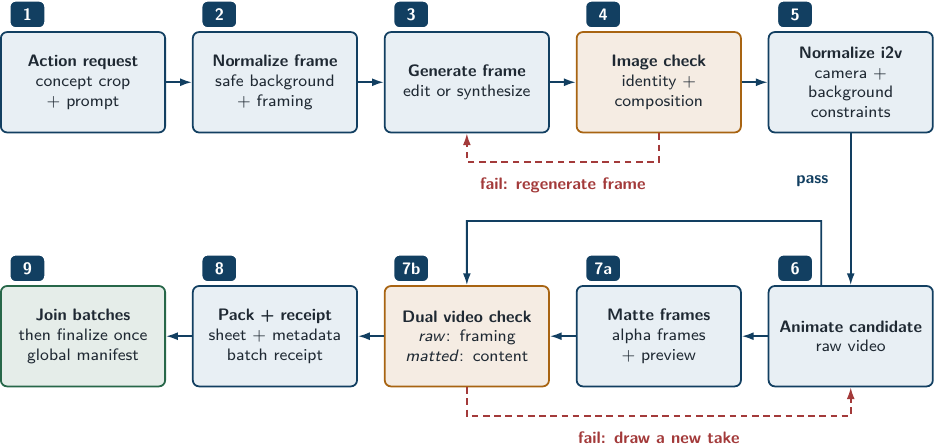}
\caption{Animated-sprite production alternates bounded generation, checks,
retries, and one post-join finalization.}
\label{fig:assetchain}
\end{figure}

\paragraph{Asset Production.}
In parallel, independent asset batches are partitioned across asset agents.
Workers write owned media and metadata sidecars but cannot touch the shared
manifest, and it is that disjoint ownership which makes parallel generation
safe. \emph{Produces}: media files, metadata sidecars, and one receipt per
batch. \Cref{fig:assetchain} shows the bounded mechanism for animated sprites.

Each entity first receives a shared visual reference. Action-specific
generation then works from a grounded concept crop, under fixed framing and
background constraints. Raw animation and matted output are checked for framing
and shipped-content quality. A failed check may trigger another animation
attempt within the configured budget; exhaustion is recorded as best effort
rather than relabeled as success. Accepted frames are packed into a Phaser
spritesheet with metadata.

\paragraph{Integration.}
The implementation workflow and every declared asset batch join before one
reducer merges file-backed sidecars into the manifest. The reducer preserves
missing required outputs as failures and records valid no-file sources, such as
procedural graphics or live text, explicitly. The asset-integration agent then
runs alone, wiring preloads, animations, scaling, display parameters, and
interface bindings. \emph{Produces}: the assembled runtime, and a sealed
receipt fixing the build digest that enters verification. The receipt is what
keeps multimodal GDD evidence distinct from the runtime assets and code that
realize it.

\begin{figure}[!ht]
\centering
\includegraphics[width=0.95\linewidth]{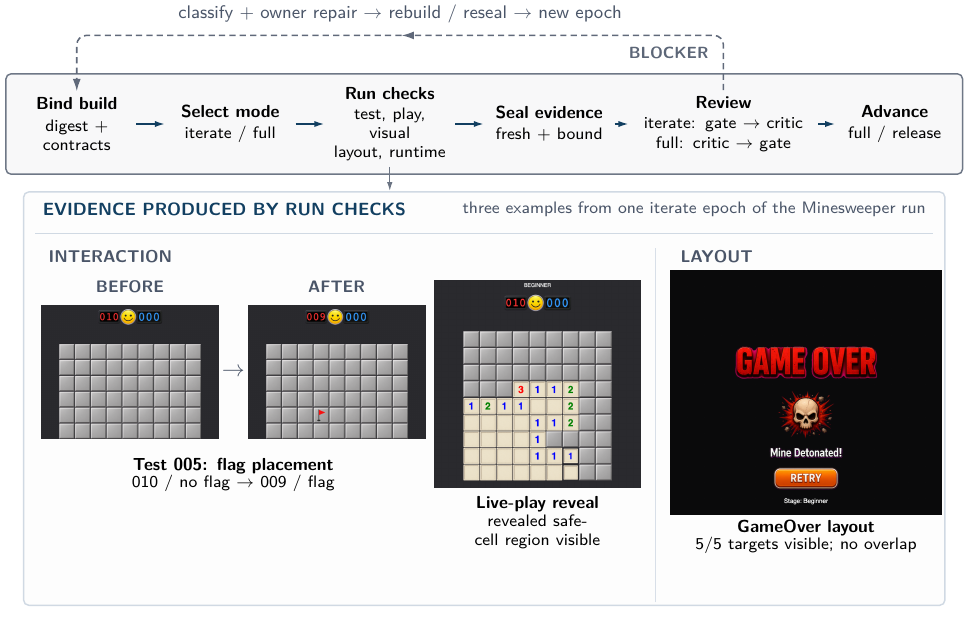}
\caption{One verification epoch. Checks run against the freshly sealed build
and produce the evidence shown here, from a Minesweeper run. The gate and the
critic then read the sealed report and either advance the run or return a
blocker to its owner for a new epoch.}
\label{fig:verification-pipeline}
\end{figure}

\paragraph{Verification.}
Verification runs on the build the previous stage sealed, and combines
deterministic and perceptual evidence without treating the two as
interchangeable. Static gates inspect files and contracts; Playwright tests
execute interaction rules~\citep{playwright2025}; a play agent explores the live
game from rendered observations and engine
state~\citep{yao2023react,he2024webvoyager}; layout checks compute bounds and
overlap; visual comparisons judge crop-paired
appearance~\citep{ku2024viescore}; and a code-blind critic assesses the
assembled evidence~\citep{zheng2023judging,gou2024critic}. \emph{Produces}: a
sealed verification report and a gate decision.
\Cref{fig:verification-pipeline} presents the flow together with one
Minesweeper iterate epoch.

Every epoch is bound to the current run, build digest, URL, contracts, and
verification identity. In \cfgkey{iterate} mode, browser, play, visual, layout,
and runtime-asset producers checkpoint their outputs, a finalizer seals the
aggregate even if a producer fails, and the iteration gate is evaluated before
the mandatory critic. In \cfgkey{full} mode, isolated recordings, video
analysis, and performance samples are added, and the final critic precedes the
release gate. Both scopes materialize deterministic records before atomic
report finalization, and neither accepts evidence from an earlier digest. The
Minesweeper epoch of \cref{fig:verification-pipeline} shows this in one run:
formal test 005 checks that a real right-click places a flag and decrements the
counter, live play checks a safe-cell reveal, and layout capture finds all five
GameOverScene targets without overlap or off-screen placement. All three are
internal production artifacts, unrelated to the frames and scores of the
external experiment.

\paragraph{Owner-Routed Repair.}
When either gate or critic reports a blocker, we group findings by owning agent
and route design repairs first~\citep{shinn2023reflexion,madaan2023selfrefine},
because they invalidate downstream projections.
Code and asset repairs are parallelized only for disjoint write sets; the
writers then join, affected shared artifacts are rebuilt, and a new build
digest is sealed before another epoch starts. \emph{Produces}: a repaired build
under a new digest, or, at budget exhaustion, a preserved record of the
unresolved findings.

\paragraph{Release.}
A game ships only once a full verification pass and the final critic clear the
release gate under a single build digest. \emph{Produces}: the released game
together with the sealed evidence that authorized it. This loop decides whether
\gca{} may advance; it does not determine the external scores in
\cref{sec:evaluation}.

\section{MCP Control Plane}
\label{sec:controlplane}

Every action that makes something or checks something goes through one MCP
server~\citep{anthropic2024mcp}. It is the one place that decides which
workspace a call belongs to, whether an artifact has the fields it is supposed
to have, and whether the evidence a gate will ask for is already there. No
agent has to write those checks itself, and no agent can skip them.
\Cref{tab:tools} groups the 55 tools by category; \cref{app:tools,app:gates}
give the full catalog and the exact thresholds.

\begin{table}[!ht]
\centering
\small
\setlength{\tabcolsep}{5pt}
\renewcommand{\arraystretch}{1.15}
\begin{tabularx}{\textwidth}{@{}lr>{\raggedright\arraybackslash}X@{}}
\toprule
\textbf{Category} & \textbf{Tools} & \textbf{Scope, with a representative tool} \\
\midrule
Lifecycle & 9 &
Workspace leasing, preflight readiness, scaffold reproject receipts, and the
transactional state journal --- \toolname{write\_pipeline\_state} \\

Asset generation and validation & 20 &
Image and video synthesis, matting, spritesheet packing, and the closed-loop
producers that retry against VLM rubrics ---
\toolname{produce\_validated\_sprite\_action} \\

Verification & 16 &
The checkpointed verification DAG, its individual producers, and the gate
runner --- \toolname{run\_verify\_dag} \\

Concept grounding & 3 &
Detect element boxes on an approved mockup and crop them into art seeds ---
\toolname{detect\_layout} \\

Post-processing and matting & 3 &
CPU compositing plus the two matting backends, remote and deterministic ---
\toolname{rmbg2\_matting} \\

Video understanding and recording & 3 &
Record a running game with sampled engine state, then analyze only that sealed
recording --- \toolname{record\_with\_state} \\

Path and layout & 1 &
Render a tower-defense path corridor from the level geometry ---
\toolname{render\_layout\_mask} \\
\midrule
\textbf{Total} & \textbf{55} & \\
\bottomrule
\end{tabularx}
\caption{MCP control-plane tools by category.}
\label{tab:tools}
\end{table}

\paragraph{Typed Tools.}
Each tool states what it takes in and what it gives back, right next to the
code that runs it and the workspace it needs. Provider calls, browser actions,
asset transformations, and gates are all declared the same way. This pays off
when several agents run at once: a bad path, a missing identity, or incomplete
evidence is refused once, at the call, instead of each agent that runs into it
deciding for itself what to make of it.

\paragraph{Workspace Leases.}
Before a tool can touch a workspace, the server pins the directory and takes an
exclusive lease on it that other processes have to respect. Two pipelines
cannot interleave writes into one game, and nobody can swap the directory out
from under a call the server has already accepted. The lease is also where the
paper trail starts: everything written after it carries the pipeline run, the
build digest, and, where it applies, the verification epoch.

\paragraph{Stamped Artifacts.}
Every result that carries evidence is stamped with the run, build, and epoch it
came from. That stamp is how a gate tells a current result from an old file
that still looks fine. A report can be complete and internally consistent and
still be thrown out, because the run and build it names are not the ones under
review.

\paragraph{Sealed Receipts.}
When one agent hands work to the next, the server writes the receipt, not the
agent: scaffold reprojection records which paths it invalidated, manifest
reduction records which batches it took in, and asset integration records which
build it wired. An agent cannot swap one of these receipts for a summary of its
own.

\paragraph{Deterministic Checks and Model Judgments.}
Some checks the server can compute; others it can only ask a model to judge,
and it keeps the two apart. Bounds, overlap, dimensions, references, counters,
and matching identities are computed outright, and asking again gives the same
answer. Whether an image looks like its reference, or whether a screen looks
finished, is a judgment against a rubric, and pinning the model and its
sampling settings makes that judgment repeatable without making it a
measurement. The first kind is a derived oracle~\citep{barr2015oracle}; the
second is evidence for a person to weigh, not a fact the pipeline can assert.

\begin{figure}[!ht]
\centering
\includegraphics[width=\linewidth]{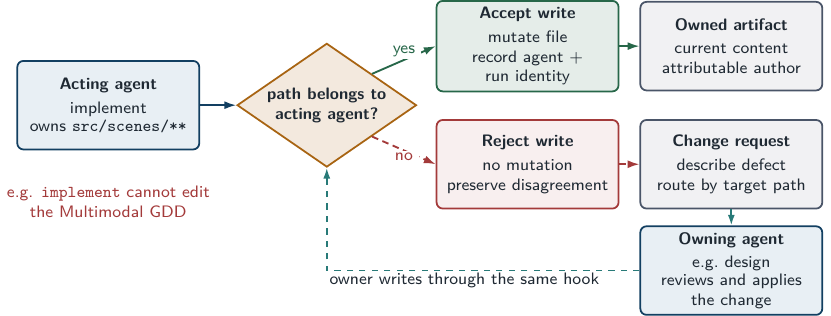}
\caption{Single-writer mutation control accepts owned writes and routes
rejected cross-component changes to the declared owner.}
\label{fig:ownership}
\end{figure}

\paragraph{Single-Writer Ownership.}
The server also gives every artifact exactly one writer, so a quick fix cannot
quietly take over someone else's file. On each write, the boundary hook
compares the path against the scope the acting agent declared, across design,
levels, scaffolding, scene code, assets, integration, and verification
evidence. \Cref{fig:ownership} shows the three outcomes: a write the agent owns
goes through, a write into another agent's file is stopped before the file
changes, and the request is handed to the agent that does own it.

\paragraph{Recovery.}
Restarting after an interruption leans on the same stamps. The orchestrator
goes back to the most recent stage boundary whose artifacts, receipts,
freshness, and ownership all still hold up, and resumes from there. A leftover
marker file does not earn credit for the work it was meant to mark, and an
output that is well-formed but stale is refused like any other. Because the
server runs these checks on every call, a run that was cut off picks up again
without mistaking half-finished writes for finished work.

Typed tools, leases, stamped outputs, receipts, and one writer per artifact are
what make a run auditable afterwards. The gates read that evidence to decide
what happens next inside the pipeline.

\section{Experiment}
\label{sec:evaluation}

We evaluate finished games independently of \gca{}'s production controls. Two
evaluations run on the same corpus: an automated one that asks how faithfully
each system realizes observable prompt requirements, and a human study that
rates the same five dimensions and records pairwise preference. The
automated evaluator receives only the original prompt and one game,
\gcaid{Evaluator(prompt, game) $\rightarrow$ scores + evidence}. It receives
neither a reference implementation nor \gca{}'s multimodal GDD, generation
trace, gate reports, or critic verdict. The protocol follows the
interaction-grounded principle of
GameCraft-Bench~\citep{luo2026gamecraft} and adapts it to Phaser browser games
and the OpenGame baseline~\citep{jiang2026opengame}.

\subsection{Setup}

\paragraph{Corpus.}
The task pool comprises 32 game briefs carrying 795 numbered MUST
requirements. A task enters the experiment when two conditions hold:
prompt-matched \gca{} and OpenGame artifacts exist for it, and a reviewed
task-specific browser suite covers its requirements. Twenty tasks satisfy both.
\Cref{tab:eval-corpus} lists this corpus under report-local sequence numbers
1--20; canonical directory identifiers and source revisions remain in the run
manifests. The corpus is fixed before scoring and is not revised afterwards.

\begin{table}[!ht]
\centering
\small
\setlength{\tabcolsep}{10pt}
\renewcommand{\arraystretch}{1.15}
\begin{tabular}{@{}rll@{\hspace{2.2cm}}rll@{}}
\toprule
\textbf{No.} & \textbf{Task} & \textbf{Family} &
\textbf{No.} & \textbf{Task} & \textbf{Family} \\
\midrule
1 & Minesweeper & Puzzle & 11 & Blackjack & Card \\
2 & Snake & Arcade & 12 & Endless Runner & Runner \\
3 & Tap Flyer & Arcade & 13 & Fruit Slicer & Action \\
4 & Shmup & Action & 14 & Solitaire (Klondike) & Card \\
5 & Match-3 & Puzzle & 15 & Link Match & Puzzle \\
6 & Tower Defense & Strategy & 16 & Gold Miner & Arcade \\
7 & Lemonade Stand & Sim. & 17 & Basketball Shoot & Sports \\
8 & Platformer & Platform & 18 & Dawn Dusk Farm & Sim. \\
9 & Deckbuilder & Card & 19 & Veinworks & Idle \\
10 & Mole Tap & Arcade & 20 & Vibe Fighter & Fighting \\
\bottomrule
\end{tabular}
\caption{Fixed 20-task paired corpus with game families, numbered in
report-local sequence. Tasks 1--10 are listed on the left and 11--20 on the
right. Task names describe gameplay genres and mechanics; they do not refer to
any commercial game.}
\label{tab:eval-corpus}
\end{table}

\paragraph{Games and scoring runs.}
Each system contributes one game per task, so the experiment compares
\(20\times2=40\) generated artifacts. Every artifact is then scored three times
under the same protocol, giving \(40\times3=120\) selected evaluation runs, 60
per system. The repeats re-score one fixed build rather than regenerate it:
deterministic checks return the same verdict on every repeat, and the spread
between repeats comes from the model-judged dimensions and from non-determinism
in the game itself~\citep{luo2014flaky}. Every selected run completed with full
task-specific requirement coverage.

\paragraph{Controlled factors.}
Prompt bytes, suite version, browser configuration, evidence policy, judge, and
retry rules are held constant across systems and repeats, so the system under
test is the only factor that varies. The fixed settings are
Chromium~\citep{chromium2026} at a \(1280\times720\) viewport under a
300-second browser ceiling; a full WebM~\citep{webm2026} recording, semantic
key frames, and 12 uniform video samples as retained evidence; and
GPT-5.6~\citep{openai2026gpt56} over two passes under a 600-second budget, with
at most three attempts per pass for connection failures, rate limits, and
retryable server errors. Every score is attributable to the SHA-256 of its
prompt, rubric, game tree, and test suite, and credentials are read at runtime
and never written into run artifacts.

\paragraph{Exclusions.}
We distinguish a failed run from an invalid measurement. A failed selected run
stays in the experiment and is scored as such, whereas an attempt whose
measurement is unsound is excluded explicitly rather than recorded as a game
failure. Six earlier Solitaire attempts fall in the second category: an
evaluator parsing defect invalidated them and valid runs superseded them, so
they remain on record as formal exclusions outside the 120 selected evaluation
runs.

\subsection{Automated Evaluation}

\paragraph{Rubric.}
Every prompt MUST requirement is preserved as an individual rubric item
assigned to Core Mechanics, Content Depth, or Functional Visuals. The harness
neither discards nor merges requirements, and it appends three task-independent
Art \& Presentation items. Missing identifiers, duplicate items, invalid
values, or incomplete coverage invalidate the corresponding judge output.
\begin{table}[!b]
\centering
\small
\setlength{\tabcolsep}{5pt}
\renewcommand{\arraystretch}{1.4}
\begin{tabularx}{\textwidth}{@{}
  >{\raggedright\arraybackslash}p{3.1cm}
  >{\raggedright\arraybackslash}p{1.7cm}
  >{\raggedright\arraybackslash}X
  >{\raggedright\arraybackslash}p{5.0cm}@{}}
\toprule
\textbf{Metric} & \textbf{Output} & \textbf{Rubric question} &
\textbf{Scoring} \\
\midrule
Requirement Satisfaction & $[0,1]$ &
What fraction of the observable prompt MUST requirements is satisfied? &
Mean of all prompt-derived Core Mechanics, Content Depth, and Functional
Visuals items. \\
Core Mechanics & $[0,1]$ &
Do controls, rules, state transitions, outcomes, and restart behavior work? &
Weighted fraction of passed deterministic MUST-item checks. \\
Content Depth & $[0,1]$ &
Are requested modes, levels, progression, and terminal content present? &
Weighted fraction of passed deterministic MUST-item checks. \\
Functional Visuals & $[0,1]$ &
Are required game state, HUD, action feedback, and outcomes visible? &
Mean of direct evidence-grounded GPT-5.6 MUST-item scores. \\
Art \& Presentation & $[0,1]$ &
Is the presentation coherent, readable, and visibly finished? &
Mean of three shared GPT-5.6 visual-quality items. \\
\bottomrule
\end{tabularx}
\caption{Test metrics and scoring rubric.}
\label{tab:eval-rubric}
\end{table}

\paragraph{Evidence.}
The evaluator serves each artifact and drives it with Playwright, delivering
every action through the rendered interface: clicks, right-clicks, hovers,
keyboard input, and drag gestures. Source, Document Object Model, layout, and
public runtime state may be read to locate rendered targets and assert
outcomes, but the suite may not invoke an internal action or mutate state in
place of user input. Tests cover controls, rules, progression, terminal states,
restart behavior, and task-specific requirements while retaining console, page,
and request errors. Each run stores scenario screenshots and a complete
gameplay recording; semantic frames are associated with named scenarios and are
not assumed to form a single chronology when scenarios initialize
independently. GPT-5.6 receives the prompt, rubric, deterministic assertion
record, key screenshots, and time-ordered frames sampled from the recording,
and an initial judgment is followed by an audit pass that rechecks every
requirement, failed assertion, terminal condition, and cited visual
observation.

\begin{table}[!b]
\centering
\small
\setlength{\tabcolsep}{6pt}
\renewcommand{\arraystretch}{1.25}
\begin{tabularx}{\textwidth}{@{}>{\raggedright\arraybackslash}X
  >{\columncolor{oursband}\centering\arraybackslash}p{3.3cm}
  >{\centering\arraybackslash}p{3.3cm}@{}}
\toprule
& \multicolumn{2}{c}{\textbf{Mean\,$\pm$\,standard deviation}} \\
\cmidrule(l){2-3}
\textbf{Metric} & \textbf{\gca{}} & \textbf{OpenGame} \\
\midrule
Requirement Satisfaction & \textbf{0.786\,$\pm$\,0.151} & 0.656\,$\pm$\,0.247 \\
Core Mechanics           & \textbf{0.852\,$\pm$\,0.144} & 0.655\,$\pm$\,0.267 \\
Content Depth            & \textbf{0.712\,$\pm$\,0.286} & 0.651\,$\pm$\,0.267 \\
Functional Visuals       & \textbf{0.711\,$\pm$\,0.192} & 0.667\,$\pm$\,0.248 \\
Art \& Presentation      & 0.797\,$\pm$\,0.099 & \textbf{0.827\,$\pm$\,0.068} \\
\bottomrule
\end{tabularx}
\caption{Aggregate prompt-grounded results. Each system contributes one game
per task and every game is scored three times, so a cell averages 60 evaluation
runs over 20 games and its deviation spans tasks as well as scoring repeats.
Bold marks the higher mean in each row. Core Mechanics and Content Depth derive
from deterministic checks,
Functional Visuals and Art \& Presentation from grounded GPT-5.6 judgments, and
Requirement Satisfaction averages the prompt-derived items across the first
three dimensions. Null values are excluded, not converted to zero.}
\label{tab:eval-results}
\end{table}

\paragraph{Scoring.}
\Cref{tab:eval-rubric} defines the five reported quantities. All item scores
lie on a continuous \([0,1]\) scale, and missing values remain unmeasured
rather than becoming zero. Core Mechanics and Content Depth are weighted
fractions of deterministic real-UI assertions that pass. Functional Visuals
instead asks whether required state, interface values, action feedback, and
outcomes are visible in rendered evidence: under the grounded-visual scoring
policy, GPT-5.6 assigns each prompt-derived item directly from screenshots and
video frames while receiving deterministic assertions as factual context. A
failed assertion must be acknowledged in the rationale and contradiction
provenance, but the assertion score is not a second numeric operand, which
preserves the distinction between an exact behavioral check and an
evidence-grounded visual judgment. GPT-5.6 scores Art \& Presentation from the
same visual evidence using three shared items: visual coherence, composition
and readability, and authored finish. Requirement Satisfaction is the mean of
all prompt-derived Core Mechanics, Content Depth, and Functional Visuals items;
Art \& Presentation is excluded because it measures general presentation rather
than a prompt MUST requirement. We compute no aggregate overall quality score.

\paragraph{Results.}
We report a scoring-only derivation from immutable earlier run manifests,
browser evidence, deterministic records, and stored GPT-5.6 item judgments. It
made no new model request, and versioned records hash-link their source
artifacts. \Cref{tab:eval-results} gives the aggregate figures and the rules
used to form them. \gca{} has higher mean scores on Requirement Satisfaction
(0.786 versus 0.656), Core Mechanics (0.852 versus 0.655), Content Depth
(0.712 versus 0.651), and Functional Visuals (0.711 versus 0.667), while
OpenGame scores higher on Art \& Presentation (0.827 versus 0.797). Each value
is computed over the 60 evaluation runs of a system's 20 games under a common
protocol, so its spread reflects differences between tasks as well as between
scoring repeats of one game. Task-level directions vary, particularly for
Content Depth and Functional Visuals.

\paragraph{Audit and caveats.}
Relative to the earlier merged scoring, the derivation changed 78 Functional
Visuals item scores across 41 runs and audited the 254 items tied to at least
one failed assertion. Its source judgments predate a stricter frame-citation
contract: 569 of 600 visual items cite an exact frame label or a child check of
a frame-labeled capture, while 31 cite scenario or assertion records even
though they received the complete frame bundle. Current judgments require an
exact rendered-frame reference for every Functional Visuals and Art \&
Presentation item. The caveat therefore concerns how those 31 items cite their
evidence; corpus coverage is unaffected.

\subsection{Human Study}

\paragraph{Protocol.}
Ten participants each evaluated all 20 prompt-matched pairs, yielding 200
participant--game ratings per system and 200 pairwise preference choices.
System identity was hidden behind randomized Game A/Game B labels, and
presentation order was counterbalanced. For each pair, participants rated both
games on anchored five-level versions of Requirement Satisfaction, Core
Mechanics, Content Depth, Functional Visuals, and Art \& Presentation, then
selected Game A, Game B, or no preference. Preference remained categorical and
was not converted to the rating scale. \Cref{app:human-questionnaire} provides
the one-page form completed once for each of a participant's 20 pairs.

\begin{table}[!b]
\centering
\small
\setlength{\tabcolsep}{6pt}
\renewcommand{\arraystretch}{1.25}
\begin{tabularx}{\textwidth}{@{}>{\raggedright\arraybackslash}X
  >{\columncolor{oursband}\centering\arraybackslash}p{3.3cm}
  >{\centering\arraybackslash}p{3.3cm}@{}}
\toprule
\textbf{Reported quantity} & \textbf{\gca{}} & \textbf{OpenGame} \\
\midrule
\emph{Rubric ratings on a 1--5 scale
  (mean\,$\pm$\,standard deviation)} & & \\
Requirement Satisfaction & \textbf{3.55\,$\pm$\,1.36} & 3.28\,$\pm$\,1.47 \\
Core Mechanics           & \textbf{3.40\,$\pm$\,1.31} & 3.16\,$\pm$\,1.36 \\
Content Depth            & \textbf{3.43\,$\pm$\,1.36} & 3.28\,$\pm$\,1.40 \\
Functional Visuals       & \textbf{3.33\,$\pm$\,1.35} & 3.29\,$\pm$\,1.33 \\
Art \& Presentation      & \textbf{3.57\,$\pm$\,1.13} & 3.03\,$\pm$\,1.19 \\
Overall                  & \textbf{3.45\,$\pm$\,1.13} & 3.21\,$\pm$\,1.18 \\
\addlinespace[3pt]
\emph{Pairwise preference, share of choices (\%)} & & \\
All 200 choices          & \textbf{43.0} & 27.5 \\
Excluding ties           & \textbf{61.0} & 39.0 \\
No preference            & \multicolumn{2}{c}{29.5} \\
\bottomrule
\end{tabularx}
\caption{Human study on 20 prompt pairs. Ten participants each rated all 20
pairs, giving 200 participant--game ratings per system. Bold marks the higher
value in each row, and \emph{Overall} is the mean of the five dimensions. Of
the 200 preference choices 59 were ``no preference''; excluding ties, \gca{} is
preferred in 86 of 141 decisive choices (two-sided sign test, $p=0.011$).}
\label{tab:human-results}
\end{table}

\paragraph{Results.}
\Cref{tab:human-results} gives the rubric means with their standard deviations
and the preference shares. \gca{} received higher mean ratings on all five
dimensions, with the largest gap in Art \& Presentation (3.57 versus 3.03).
Participants preferred \gca{} in 86 of 200 pair choices, OpenGame in 55, and
expressed no preference in 59. Among the 141 decisive choices, the preference
split was 61.0\% versus 39.0\% in favor of \gca{} (two-sided sign test,
\(p=0.011\)). The ratings complement the automated results without standing in
for their prompt-derived, interaction-grounded evidence.

\subsection{Case Study: Gold Miner}
\label{sec:case-study}

\begin{figure}[!ht]
\centering
\includegraphics[width=\linewidth]{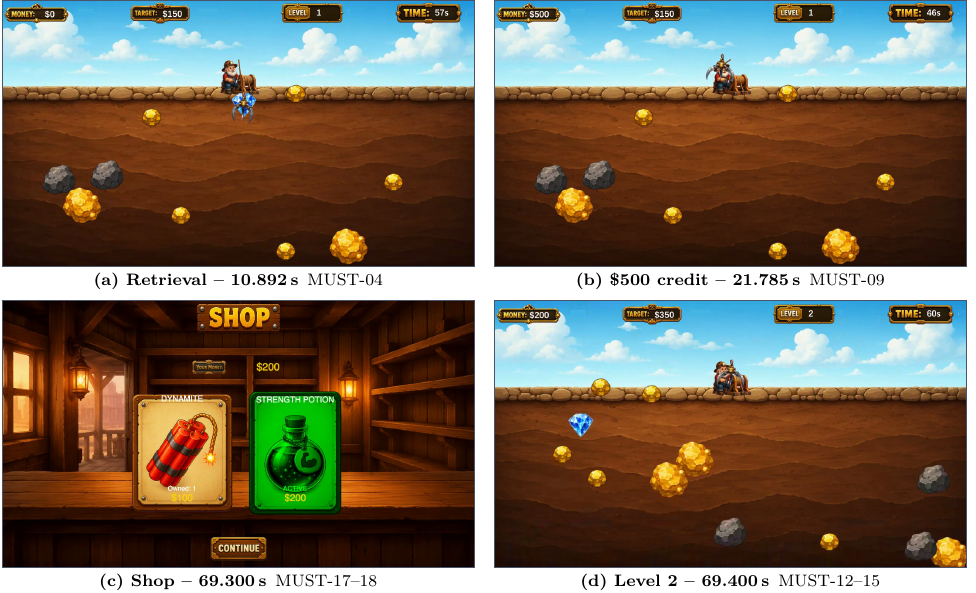}
\caption{Gold Miner formal-run evidence: retrieval, exact \$500 credit, shop
purchase, and level-2 transition in one continuous recording.}
\label{fig:case-study-timeline}
\end{figure}

We use Gold Miner to connect the architecture to a formal external run while
preserving the evidence boundary. Its prompt asks for claw retrieval, mineral
values, a countdown timer, cumulative money, per-level targets, shop items,
runtime states, and interface readouts. The multimodal GDD maps those
requirements to controls, transitions, scenes, and screen concepts, and
scaffold projects the scene, input, objective, layout, and observability
contracts.

The internal and external records disagree about this run, and that is the
point of reporting it. The pipeline ended blocked rather than released: a
timing-sensitive launch guard remained, visual-comparison provider calls
failed, and metadata checks were unresolved. The external suite, driving the
rendered interface with runtime state read only for assertions, observed a
coherent prefix of the requested game.
\Cref{fig:case-study-timeline} shows four frames from one continuous recording,
covering retrieval, the exact credit after collection and removal, a shop
purchase, and the transition into the next level with the balance carried over.
Support is partial for claw-tip collision, final victory, and fresh restart,
and collection is inconsistent across the other scoring repeats.

Neither record subsumes the other. The gates withheld release on evidence the
external suite never sees, while the external suite confirmed prompt
fulfillment that a blocked internal status would otherwise hide. Reading the
two separately is what keeps each of them meaningful.

\FloatBarrier

\section{Conclusion}
\label{sec:conclusion}

We presented \gca{}, a pipeline that turns one natural-language request into a
complete, playable Phaser browser game: a multimodal GDD grounds design intent
in scene evidence, scaffold projects that document into a runtime shell with
machine-readable contracts and deterministic tests, eight specialized agents
implement scenes serially while generating assets in parallel, and a typed MCP
control plane binds every artifact, receipt, and item of evidence to one run
and build digest. To decide whether a game is done, \gca{} tests the build it
just assembled instead of trusting what an agent says it did. That is why it
can run with nobody watching and still refuse to ship a game that only looks
finished. Judged with rendered interfaces, the games it produces
satisfy prompt requirements and implement working mechanics at a level
comparable open pipelines do not reach.

\section*{Contributors}

\textbf{Core contributors.}
Xiangxin~Zhou,
Jiaqi~Tang,
Ke~Ma,
Wenhao~Yang,
Jiarui~Yao,
Boye~Niu,
Haiyang~Guo,
Yifei~Zhao,
Huanjin~Yao,
Tianyu~Pang.

\textbf{Other contributors.}
Pu~Jian,
Ruihuang~Li,
Haitao~Wu,
Yushi~Huang,
Xiangyan~Liu,
Yan~Sun,
Kaixin~Li.

\clearpage
\bibliography{ref}

@inproceedings{qian2024chatdev,
  title     = {{ChatDev}: Communicative Agents for Software Development},
  author    = {Qian, Chen and Liu, Wei and Liu, Hongzhang and Chen, Nuo and
               Dang, Yufan and Li, Jiahao and Yang, Cheng and Chen, Weize and
               Su, Yusheng and Cong, Xin and Xu, Juyuan and Li, Dahai and
               Liu, Zhiyuan and Sun, Maosong},
  booktitle = {Proceedings of the 62nd Annual Meeting of the Association for
               Computational Linguistics (ACL)},
  year      = {2024}
}

@inproceedings{hong2024metagpt,
  title     = {{MetaGPT}: Meta Programming for a Multi-Agent Collaborative
               Framework},
  author    = {Hong, Sirui and Zhuge, Mingchen and Chen, Jonathan and
               Zheng, Xiawu and Cheng, Yuheng and Zhang, Ceyao and
               Wang, Jinlin and Wang, Zili and Yau, Steven Ka Shing and
               Lin, Zijuan and Zhou, Liyang and Ran, Chenyu and
               Xiao, Lingfeng and Wu, Chenglin and Schmidhuber, J{\"u}rgen},
  booktitle = {International Conference on Learning Representations (ICLR)},
  year      = {2024}
}

@inproceedings{wu2024autogen,
  title     = {{AutoGen}: Enabling Next-Gen {LLM} Applications via Multi-Agent
               Conversation},
  author    = {Wu, Qingyun and Bansal, Gagan and Zhang, Jieyu and Wu, Yiran and
               Li, Beibin and Zhu, Erkang and Jiang, Li and Zhang, Xiaoyun and
               Zhang, Shaokun and Liu, Jiale and Awadallah, Ahmed Hassan and
               White, Ryen W and Burger, Doug and Wang, Chi},
  booktitle = {Conference on Language Modeling (COLM)},
  year      = {2024}
}

@inproceedings{jimenez2024swebench,
  title     = {{SWE-bench}: Can Language Models Resolve Real-World GitHub
               Issues?},
  author    = {Jimenez, Carlos E and Yang, John and Wettig, Alexander and
               Yao, Shunyu and Pei, Kexin and Press, Ofir and
               Narasimhan, Karthik},
  booktitle = {International Conference on Learning Representations (ICLR)},
  year      = {2024}
}

@article{chen2021codex,
  title   = {Evaluating Large Language Models Trained on Code},
  author  = {Chen, Mark and Tworek, Jerry and Jun, Heewoo and Yuan, Qiming and
             Pinto, Henrique Ponde de Oliveira and Kaplan, Jared and
             Edwards, Harri and Burda, Yuri and Joseph, Nicholas and
             Brockman, Greg and others},
  journal = {arXiv preprint arXiv:2107.03374},
  year    = {2021}
}

@article{austin2021program,
  title   = {Program Synthesis with Large Language Models},
  author  = {Austin, Jacob and Odena, Augustus and Nye, Maxwell and
             Bosma, Maarten and Michalewski, Henryk and Dohan, David and
             Jiang, Ellen and Cai, Carrie and Terry, Michael and Le, Quoc and
             Sutton, Charles},
  journal = {arXiv preprint arXiv:2108.07732},
  year    = {2021}
}

@article{barr2015oracle,
  title   = {The Oracle Problem in Software Testing: A Survey},
  author  = {Barr, Earl T and Harman, Mark and McMinn, Phil and
             Shahbaz, Muzammil and Yoo, Shin},
  journal = {IEEE Transactions on Software Engineering},
  volume  = {41},
  number  = {5},
  pages   = {507--525},
  year    = {2015}
}

@inproceedings{luo2014flaky,
  title     = {An Empirical Analysis of Flaky Tests},
  author    = {Luo, Qingzhou and Hariri, Farah and Eloussi, Lamyaa and
               Marinov, Darko},
  booktitle = {Proceedings of the 22nd ACM SIGSOFT International Symposium on
               Foundations of Software Engineering (FSE)},
  pages     = {643--653},
  year      = {2014}
}

@inproceedings{yao2023react,
  title     = {{ReAct}: Synergizing Reasoning and Acting in Language Models},
  author    = {Yao, Shunyu and Zhao, Jeffrey and Yu, Dian and Du, Nan and
               Shafran, Izhak and Narasimhan, Karthik and Cao, Yuan},
  booktitle = {International Conference on Learning Representations (ICLR)},
  year      = {2023}
}

@inproceedings{huang2024selfcorrect,
  title     = {Large Language Models Cannot Self-Correct Reasoning Yet},
  author    = {Huang, Jie and Chen, Xinyun and Mishra, Swaroop and
               Zheng, Huaixiu Steven and Yu, Adams Wei and Song, Xinying and
               Zhou, Denny},
  booktitle = {International Conference on Learning Representations (ICLR)},
  year      = {2024}
}

@inproceedings{gou2024critic,
  title     = {{CRITIC}: Large Language Models Can Self-Correct with
               Tool-Interactive Critiquing},
  author    = {Gou, Zhibin and Shao, Zhihong and Gong, Yeyun and
               Shen, Yelong and Yang, Yujiu and Duan, Nan and Chen, Weizhu},
  booktitle = {International Conference on Learning Representations (ICLR)},
  year      = {2024}
}

@inproceedings{shinn2023reflexion,
  title     = {Reflexion: Language Agents with Verbal Reinforcement Learning},
  author    = {Shinn, Noah and Cassano, Federico and Berman, Edward and
               Gopinath, Ashwin and Narasimhan, Karthik and Yao, Shunyu},
  booktitle = {Advances in Neural Information Processing Systems (NeurIPS)},
  year      = {2023}
}

@inproceedings{madaan2023selfrefine,
  title     = {Self-Refine: Iterative Refinement with Self-Feedback},
  author    = {Madaan, Aman and Tandon, Niket and Gupta, Prakhar and
               Hallinan, Skyler and Gao, Luyu and Wiegreffe, Sarah and
               Alon, Uri and Dziri, Nouha and Prabhumoye, Shrimai and
               Yang, Yiming and Gupta, Shashank and
               Majumder, Bodhisattwa Prasad and Hermann, Katherine and
               Welleck, Sean and Yazdanbakhsh, Amir and Clark, Peter},
  booktitle = {Advances in Neural Information Processing Systems (NeurIPS)},
  year      = {2023}
}

@inproceedings{zheng2023judging,
  title     = {Judging {LLM}-as-a-Judge with {MT-Bench} and {Chatbot Arena}},
  author    = {Zheng, Lianmin and Chiang, Wei-Lin and Sheng, Ying and
               Zhuang, Siyuan and Wu, Zhanghao and Zhuang, Yonghao and
               Lin, Zi and Li, Zhuohan and Li, Dacheng and Xing, Eric P and
               Zhang, Hao and Gonzalez, Joseph E and Stoica, Ion},
  booktitle = {Advances in Neural Information Processing Systems (NeurIPS)
               Datasets and Benchmarks Track},
  year      = {2023}
}

@article{jiang2026opengame,
  title   = {{OpenGame}: Open Agentic Coding for Games},
  author  = {Jiang, Yilei and Hu, Jinyuan and Xiao, Qianyin and
             Zheng, Yaozhi and Ma, Ruize and Feng, Kaituo and Han, Jiaming and
             Peng, Tianshuo and Fan, Kaixuan and Zhang, Manyuan and
             Yue, Xiangyu},
  journal = {arXiv preprint arXiv:2604.18394},
  year    = {2026}
}

@article{luo2026gamecraft,
  title   = {{GameCraft-Bench}: Can Agents Build Playable Games End-to-End in a
             Real Game Engine?},
  author  = {Luo, Tongxu and Wang, Rongsheng and Bi, Jiaxi and
             Xu, Chenming and Tang, Zhengyang and others},
  journal = {arXiv preprint arXiv:2606.17861},
  year    = {2026}
}

@article{wang2025titan,
  title   = {Leveraging {LLM} Agents for Automated Video Game Testing},
  author  = {Wang, Chengjia and Tang, Lanling and Yuan, Ming and Yu, Jiongchi
             and Xie, Xiaofei and Bu, Jiajun},
  journal = {arXiv preprint arXiv:2509.22170},
  year    = {2025}
}

@article{jiang2026gbqa,
  title   = {{GBQA}: A Game Benchmark for Evaluating {LLMs} as Quality
             Assurance Engineers},
  author  = {Jiang, Shufan and Chen, Chios and Chen, Zhiyang},
  journal = {arXiv preprint arXiv:2604.02648},
  year    = {2026}
}

@article{wang2023voyager,
  title   = {Voyager: An Open-Ended Embodied Agent with Large Language Models},
  author  = {Wang, Guanzhi and Xie, Yuqi and Jiang, Yunfan and
             Mandlekar, Ajay and Xiao, Chaowei and Zhu, Yuke and Fan, Linxi
             and Anandkumar, Anima},
  journal = {arXiv preprint arXiv:2305.16291},
  year    = {2023}
}

@article{chen2023gamegpt,
  title   = {{GameGPT}: Multi-agent Collaborative Framework for Game
             Development},
  author  = {Chen, Dake and Wang, Hanbin and Huo, Yunhao and Li, Yuzhao and
             Zhang, Haoyang},
  journal = {arXiv preprint arXiv:2310.08067},
  year    = {2023}
}

@inproceedings{chi2026gamedevbench,
  title     = {{GameDevBench}: Evaluating Agentic Capabilities Through Game
               Development},
  author    = {Chi, Wayne and Fang, Yixiong and Yayavaram, Arnav and
               Yayavaram, Siddharth and Karten, Seth and Wei, Qiuhong Anna and
               Chen, Runkun and Wang, Alexander and Chen, Valerie and
               Talwalkar, Ameet and Donahue, Chris},
  booktitle = {International Conference on Machine Learning (ICML)},
  year      = {2026}
}

@inproceedings{zheng2019wuji,
  title     = {Wuji: Automatic Online Combat Game Testing Using Evolutionary
               Deep Reinforcement Learning},
  author    = {Zheng, Yan and Xie, Xiaofei and Su, Ting and Ma, Lei and
               Hao, Jianye and Meng, Zhaopeng and Liu, Yang and Shen, Ruimin
               and Chen, Yinfeng and Fan, Changjie},
  booktitle = {IEEE/ACM International Conference on Automated Software
               Engineering (ASE)},
  year      = {2019}
}

@article{zhang2025videogamebench,
  title   = {{VideoGameBench}: Can Vision-Language Models Complete Popular
             Video Games?},
  author  = {Zhang, Alex L and Griffiths, Thomas L and Narasimhan, Karthik R
             and Press, Ofir},
  journal = {arXiv preprint arXiv:2505.18134},
  year    = {2025}
}

@inproceedings{sudhakaran2023mariogpt,
  title     = {{MarioGPT}: Open-Ended Text2Level Generation through Large
               Language Models},
  author    = {Sudhakaran, Shyam and Gonz{\'a}lez-Duque, Miguel and
               Freiberger, Matthias and Glanois, Claire and Najarro, Elias and
               Risi, Sebastian},
  booktitle = {Advances in Neural Information Processing Systems (NeurIPS)},
  year      = {2023}
}

@book{shaker2016pcg,
  title     = {Procedural Content Generation in Games},
  author    = {Shaker, Noor and Togelius, Julian and Nelson, Mark J},
  publisher = {Springer},
  year      = {2016}
}

@article{summerville2018pcgml,
  title   = {Procedural Content Generation via Machine Learning ({PCGML})},
  author  = {Summerville, Adam and Snodgrass, Sam and Guzdial, Matthew and
             Holmg{\aa}rd, Christoffer and Hoover, Amy K and Isaksen, Aaron and
             Nealen, Andy and Togelius, Julian},
  journal = {IEEE Transactions on Games},
  volume  = {10},
  number  = {3},
  pages   = {257--270},
  year    = {2018}
}

@inproceedings{rombach2022ldm,
  title     = {High-Resolution Image Synthesis with Latent Diffusion Models},
  author    = {Rombach, Robin and Blattmann, Andreas and Lorenz, Dominik and
               Esser, Patrick and Ommer, Bj{\"o}rn},
  booktitle = {IEEE/CVF Conference on Computer Vision and Pattern Recognition
               (CVPR)},
  year      = {2022}
}

@article{seed2025seedance,
  title   = {Seedance 1.0: Exploring the Boundaries of Video Generation
             Models},
  author  = {{ByteDance Seed Vision Team}},
  journal = {arXiv preprint arXiv:2506.09113},
  year    = {2025}
}

@article{blattmann2023svd,
  title   = {Stable Video Diffusion: Scaling Latent Video Diffusion Models to
             Large Datasets},
  author  = {Blattmann, Andreas and Dockhorn, Tim and Kulal, Sumith and
             Mendelevitch, Daniel and Kilian, Maciej and Lorenz, Dominik and
             Levi, Yam and English, Zion and Voleti, Vikram and Letts, Adam and
             others},
  journal = {arXiv preprint arXiv:2311.15127},
  year    = {2023}
}

@inproceedings{ku2024viescore,
  title     = {{VIEScore}: Towards Explainable Metrics for Conditional Image
               Synthesis Evaluation},
  author    = {Ku, Max and Jiang, Dongfu and Wei, Cong and Yue, Xiang and
               Chen, Wenhu},
  booktitle = {Proceedings of the 62nd Annual Meeting of the Association for
               Computational Linguistics (ACL)},
  year      = {2024}
}

@inproceedings{he2024webvoyager,
  title     = {{WebVoyager}: Building an End-to-End Web Agent with Large
               Multimodal Models},
  author    = {He, Hongliang and Yao, Wenlin and Ma, Kaixin and Yu, Wenhao and
               Dai, Yong and Zhang, Hongming and Lan, Zhenzhong and Yu, Dong},
  booktitle = {Proceedings of the 62nd Annual Meeting of the Association for
               Computational Linguistics (ACL)},
  year      = {2024}
}

@inproceedings{zhou2024webarena,
  title     = {{WebArena}: A Realistic Web Environment for Building Autonomous
               Agents},
  author    = {Zhou, Shuyan and Xu, Frank F and Zhu, Hao and Zhou, Xuhui and
               Lo, Robert and Sridhar, Abishek and Cheng, Xianyi and
               Ou, Tianyue and Bisk, Yonatan and Fried, Daniel and Alon, Uri
               and Neubig, Graham},
  booktitle = {International Conference on Learning Representations (ICLR)},
  year      = {2024}
}

@inproceedings{zheng2024seeact,
  title     = {{GPT-4V(ision)} is a Generalist Web Agent, if Grounded},
  author    = {Zheng, Boyuan and Gou, Boyu and Kil, Jihyung and Sun, Huan and
               Su, Yu},
  booktitle = {International Conference on Machine Learning (ICML)},
  year      = {2024}
}

@inproceedings{cheng2024seeclick,
  title     = {{SeeClick}: Harnessing {GUI} Grounding for Advanced Visual
               {GUI} Agents},
  author    = {Cheng, Kanzhi and Sun, Qiushi and Chu, Yougang and Xu, Fangzhi and
               Li, Yantao and Zhang, Jianbing and Wu, Zhiyong},
  booktitle = {Proceedings of the 62nd Annual Meeting of the Association for
               Computational Linguistics (ACL)},
  year      = {2024}
}

@misc{anthropic2024mcp,
  title        = {Model Context Protocol},
  author       = {{Anthropic}},
  year         = {2024},
  howpublished = {\url{https://modelcontextprotocol.io}}
}

@misc{gptimage2,
    title = {{GPT-Image-2}},
    author = {{OpenAI}},
    year = {2026},
    howpublished = {\url{https://developers.openai.com/api/docs/models/gpt-image-2}}
}

@misc{briaai2024rmbg,
  title        = {{RMBG-2.0}: Background Removal Model},
  author       = {{BRIA AI}},
  year         = {2024},
  howpublished = {\url{https://huggingface.co/briaai/RMBG-2.0}}
}

@misc{phaser2025,
  title        = {Phaser: A Fast, Free and Fun Open Source Framework for
                  Canvas and {WebGL} Powered Browser Games},
  author       = {{Phaser Studio}},
  year         = {2025},
  howpublished = {\url{https://phaser.io/}}
}

@misc{playwright2025,
  title        = {Playwright: Fast and Reliable End-to-End Testing for Modern
                  Web Apps},
  author       = {{Microsoft}},
  year         = {2025},
  howpublished = {\url{https://playwright.dev/}}
}

@misc{openai2026gpt56,
  title        = {{GPT-5.6}},
  author       = {{OpenAI}},
  year         = {2026},
  howpublished = {\url{https://developers.openai.com/api/docs/models/gpt-5.6}}
}

@misc{chromium2026,
  title        = {Chromium: An Open-Source Browser Project},
  author       = {{The Chromium Projects}},
  year         = {2026},
  howpublished = {\url{https://www.chromium.org/Home/}}
}

@misc{webm2026,
  title        = {{WebM}: An Open Web Media Project},
  author       = {{The WebM Project}},
  year         = {2026},
  howpublished = {\url{https://www.webmproject.org/}}
}

@misc{godot2026,
  title        = {Godot Engine: A Free and Open Source {2D} and {3D} Game
                  Engine},
  author       = {{Godot Engine Contributors}},
  year         = {2026},
  howpublished = {\url{https://godotengine.org/}}
}

\clearpage
\appendix
\raggedbottom
The appendix is the system reference behind the main text.
Appendix~\ref{app:tools} catalogs every tool on the MCP control plane.
Appendix~\ref{app:gates} lists every gate, the membership of each composite
gate, the gate ladder, and the complete threshold reference.
Appendix~\ref{app:rubrics} gives the VLM rubrics used for asset acceptance and
release review.
Appendix~\ref{app:interfaces} specifies the agent dispatch interfaces.
Appendix~\ref{app:human-questionnaire} provides the one-page questionnaire
repeated for each of a participant's 20 assigned pairs.

\section{MCP Tool Reference}
\label{app:tools}

This is the complete control plane, enumerated from the live tool registry
instead of from documentation. All tools are exposed over the Model Context
Protocol~\citep{anthropic2024mcp}. The \textbf{W} column marks tools that
declare \cfgkey{requires\_workspace} and therefore refuse to run outside a
leased game workspace; the dispatcher enforces this centrally, so no handler
re-checks it. External backends appear with the tools that wrap them:
GPT-Image-2~\citep{gptimage2} for image synthesis,
Seedance~\citep{seed2025seedance} for animation,
RMBG-2.0~\citep{briaai2024rmbg} for matting, and
Playwright~\citep{playwright2025} for browser work. The generated game runs on
Phaser~\citep{phaser2025}.

\newlength{\trname}\setlength{\trname}{4.3cm}
\newlength{\trflag}\setlength{\trflag}{0.45cm}
\newlength{\trwhat}
\setlength{\trwhat}{\dimexpr\textwidth-\trname-\trflag-20pt\relax}

\subsection{Lifecycle (9)}

\begingroup
\small
\setlength{\tabcolsep}{5pt}
\renewcommand{\arraystretch}{1.40}
\begin{longtable}{@{}>{\raggedright\arraybackslash}p{\trname}%
  >{\centering\arraybackslash}p{\trflag}%
  >{\raggedright\arraybackslash}p{\trwhat}@{}}
\caption{Lifecycle tools.}
\label{tab:tools-lifecycle}\\
\toprule
\textbf{Tool} & \textbf{W} & \textbf{Purpose} \\
\midrule
\endfirsthead
\toprule
\textbf{Tool} & \textbf{W} & \textbf{Purpose} \\
\midrule
\endhead
\toolname{ping} & & Health check: workspace and credential status. \\
\toolname{init\_workspace} & & Open one game directory under explicit
\cfgkey{create} or \cfgkey{resume} intent; pins the inode and takes an
exclusive repository-wide lease. \\
\toolname{close\_workspace} & & Detach the workspace and release its lease;
gated tools then fail. \\
\toolname{list\_tools\_summary} & & One-line summary of every registered
tool. \\
\toolname{preflight\_check} & & Startup readiness: credentials, R2 upload
config, filesystem, Playwright, ffmpeg/ffprobe, matting endpoint. \\
\toolname{asset\_pipeline\_smoke\_test} & \cyes & End-to-end probe before asset
work: tiny image upload, matting, video credentials. \\
\toolname{seal\_scaffold\_reproject} & \cyes & Seal the scaffold-declared stale
path list into a receipt; callers cannot supply or broaden the paths. \\
\toolname{delete\_scaffold\_owned\_paths} & \cyes & Consume exactly one sealed
receipt and delete its obsolete scene, helper, and test files. \\
\toolname{write\_pipeline\_state} & \cyes & Transactional journal and write of
\artifact{state.json} and \artifact{resume.json}; preserves run identity,
enforces monotonic counters. \\
\bottomrule
\end{longtable}
\endgroup

\subsection{Asset Generation and Validation (20)}

\begingroup
\small
\setlength{\tabcolsep}{5pt}
\renewcommand{\arraystretch}{1.40}
\begin{longtable}{@{}>{\raggedright\arraybackslash}p{\trname}%
  >{\centering\arraybackslash}p{\trflag}%
  >{\raggedright\arraybackslash}p{\trwhat}@{}}
\caption{Asset-generation and validation tools.}
\label{tab:tools-assets}\\
\toprule
\textbf{Tool} & \textbf{W} & \textbf{Purpose} \\
\midrule
\endfirsthead
\toprule
\textbf{Tool} & \textbf{W} & \textbf{Purpose} \\
\midrule
\endhead
\toolname{generate\_image} & \cyes & Text-to-image; wraps the prompt in a
chroma-key background constraint for mattable asset types. \\
\toolname{edit\_image} & \cyes & Reference-conditioned edit; derives a
game-ready sprite from an approved concept crop. \\
\toolname{generate\_seedance\_video} & \cyes & Image-to-video or text-to-video
on the active provider channel. \\
\toolname{extend\_background} & \cyes & Seamless rightward outpainting for
static backgrounds. \\
\toolname{build\_i2v\_prompt} & & Wrap an animation prompt in the validated
constraint block: background lock, hard-edged effects, no glow or fade. \\
\toolname{rmbg2\_video\_matting} & \cyes & Per-frame matting to an RGBA
sequence plus encoded transparent outputs. \\
\toolname{validate\_asset\_video} & \cyes & Rubric check on an animation, run
against the raw or matted video depending on rubric group. \\
\toolname{produce\_validated\_video} & \cyes & Generate, matte, and two-pass
validate an animation, retrying within configured bounds. \\
\toolname{validate\_image} & \cyes & Rubric check on a sprite first frame; the
pre-animation gate. \\
\toolname{validate\_asset\_slot} & \cyes & Slot-fit check of a finished runtime
artifact against its design slot; batchable across the manifest. \\
\toolname{produce\_validated\_first\_frame} & \cyes & Closed
generation--validation loop, retrying on failure with validator feedback. \\
\toolname{produce\_validated\_sprite\_action} & \cyes & Full first-frame,
animation, and matte chain in one call, with per-stage validation
strategies. \\
\toolname{analyze\_sprite\_metadata} & \cyes & Content bounding box from the
alpha channel plus inferred facing. \\
\toolname{pack\_spritesheet} & \cyes & Pack matted frames into a Phaser sheet
and JSON sidecar under the texture-size limit. \\
\toolname{register\_existing\_static\_asset} & \cyes & Seal provenance for a
supplied or licensed PNG without regenerating it. \\
\toolname{finalize\_asset\_manifest} & \cyes & Single-writer post-join reducer;
re-opens every file and validates measured dimensions against the manifest. \\
\toolname{content\_bbox} & \cyes & Opaque-content bounds of a static asset for
collision and display metadata. \\
\toolname{trim\_transparent} & \cyes & Crop a static asset to its alpha content
box. \\
\toolname{scale\_uniform} & \cyes & Aspect-preserving resize to the authored
in-game display size. \\
\toolname{produce\_static\_asset} & \cyes & Recommended default for static
props and UI: generation, matting, trimming, scaling, and sealed publication in
one call. \\
\bottomrule
\end{longtable}
\endgroup

\subsection{Verification (16)}

\begingroup
\small
\setlength{\tabcolsep}{5pt}
\renewcommand{\arraystretch}{1.40}
\begin{longtable}{@{}>{\raggedright\arraybackslash}p{\trname}%
  >{\centering\arraybackslash}p{\trflag}%
  >{\raggedright\arraybackslash}p{\trwhat}@{}}
\caption{Verification tools.}
\label{tab:tools-verification}\\
\toprule
\textbf{Tool} & \textbf{W} & \textbf{Purpose} \\
\midrule
\endfirsthead
\toprule
\textbf{Tool} & \textbf{W} & \textbf{Purpose} \\
\midrule
\endhead
\toolname{run\_verify\_dag} & \cyes & Canonical pass: opens or resumes an
epoch, runs producers concurrently, fans in, finalizes in a \texttt{finally}
block, and gates. \\
\toolname{begin\_verify\_run} & \cyes & Low-level epoch open under the
workspace lock; diagnostics and recovery only. \\
\toolname{run\_tests} & \cyes & Playwright suite against the epoch URL; a
runtime token binds the browser to this workspace. \\
\toolname{play\_game} & \cyes & Closed-loop play agent scoring itself against
the executable objectives specification. \\
\toolname{visual\_diff} & \cyes & Whole-screen and per-element perceptual
comparison against concept mockups and crops. \\
\toolname{capture\_layout} & \cyes & Deterministic capture of rendered object
bounds for the layout axis. \\
\toolname{capture\_runtime\_assets} & \cyes & Capture loaded texture keys,
animations, and display-list identity from the running game. \\
\toolname{materialize\_gameplay\_alignment} & \cyes & Project the play report
onto the gameplay axis. \\
\toolname{materialize\_physics\_alignment} & \cyes & Project body, art, and
world checks onto the physics axis, or mark it not applicable. \\
\toolname{materialize\_layout\_alignment} & \cyes & Non-blocking evaluation of
the layout axis from captured bounds. \\
\toolname{write\_asset\_integration\_report} & \cyes & Seal the manifest, file,
and runtime-wiring evidence produced by integration. \\
\toolname{write\_verify\_report} & \cyes & Fan producer receipts into the
epoch's aggregate report. \\
\toolname{gate\_check} & \cyes & Run a named gate; returns \cfgkey{blocked}
plus routed \cfgkey{failures[]}. \\
\toolname{check\_ui\_contract} & \cyes & Advisory standalone UI contract
diagnostic; not a release authority. \\
\toolname{check\_td\_contract} & \cyes & Advisory tower-defense structural
diagnostic. \\
\toolname{check\_interaction\_contract} & \cyes & Advisory collision and
interaction diagnostic. \\
\bottomrule
\end{longtable}
\endgroup

\subsection{Concept Grounding, Post-Processing, Video, and Layout (10)}

\begingroup
\small
\setlength{\tabcolsep}{5pt}
\renewcommand{\arraystretch}{1.40}
\begin{longtable}{@{}>{\raggedright\arraybackslash}p{\trname}%
  >{\centering\arraybackslash}p{\trflag}%
  >{\raggedright\arraybackslash}p{\trwhat}@{}}
\caption{Concept-grounding, post-processing, video, and layout tools.}
\label{tab:tools-support}\\
\toprule
\textbf{Tool} & \textbf{W} & \textbf{Purpose} \\
\midrule
\endfirsthead
\toprule
\textbf{Tool} & \textbf{W} & \textbf{Purpose} \\
\midrule
\endhead
\multicolumn{3}{@{}l}{\emph{Concept grounding (3)}} \\
\toolname{detect\_layout} & & Precision vision pass: tight canvas-relative
boxes with roles and anchors for a mockup's elements. \\
\toolname{ground\_crop} & & Cut detected boxes into per-element reference
images that seed asset generation. \\
\toolname{fit\_image\_to\_resolution} & & Fit a concept mockup to the exact
game resolution; design files only, never runtime assets. \\
\addlinespace[3pt]
\multicolumn{3}{@{}l}{\emph{Post-processing and matting (3)}} \\
\toolname{rmbg2\_matting} & \cyes & Remote soft-alpha background removal; the
preferred backend. \\
\toolname{color\_key\_matting} & \cyes & Deterministic CPU chroma-key removal;
the outage fallback. \\
\toolname{postprocess} & \cyes & CPU utilities: layered composite previews and
seamless tiling. \\
\addlinespace[3pt]
\multicolumn{3}{@{}l}{\emph{Video understanding and recording (3)}} \\
\toolname{record\_with\_state} & \cyes & Record the running game while sampling
engine state; emits video, state timeline, performance summary, and a
receipt. \\
\toolname{analyze\_video} & \cyes & Analyze only a recording sealed by the
current epoch's receipt; rejects foreign or hash-mismatched video. \\
\toolname{get\_state\_extractor} & & Return the bundled state-extractor module
that scaffolded tests import. \\
\addlinespace[3pt]
\multicolumn{3}{@{}l}{\emph{Path and layout (1)}} \\
\toolname{render\_layout\_mask} & \cyes & Render a tower-defense path-corridor
mask from the level geometry source of truth. \\
\bottomrule
\end{longtable}
\endgroup

\clearpage
\section{Gate and Threshold Reference}
\label{app:gates}

\subsection{Composite Gate Membership}

Besides the four composites below and the \gatename{step\_done} phase
diagnostic, every registry name is an individual check, callable standalone for
diagnosis; \gatename{levels} fans out to its own per-level checks, and three
emit JSON rather than a verdict (\gatename{build\_digest},
\gatename{gate\_contract}, \gatename{materialize\_layout\_alignment}).

\gatename{scaffold\_done} runs 7 checks: \gatename{basic\_structure},
\gatename{pipeline\_state}, \gatename{levels}, \gatename{scaffold\_contracts},
\gatename{test\_contract}, \gatename{rubric\_contract}, and
\gatename{asset\_manifest}. It is not a subset of the release chain:
\gatename{pipeline\_state} and \gatename{scaffold\_contracts} appear in no other
composite, because the projected contracts they guard are superseded once
implementation begins.

\gatename{iteration\_ready} runs the 32 checks in \cref{tab:iterchecks} with the
verification report restricted to \cfgkey{iterate} mode; genre-specific checks
return not applicable outside their genre. \gatename{first\_pass\_complete} runs
the same 32 under \cfgkey{full} or \cfgkey{full-novideo} and adds
\gatename{release\_waivers}, \gatename{playthrough\_recorded}, and
\gatename{performance}, with 35 checks in total. \gatename{ready\_to\_ship} adds
\gatename{no\_open\_issues}, \gatename{critic}, \gatename{polish\_rounds}, and
\gatename{play\_quality}, for 39 checks; \gatename{gate\_contract} machine-checks the
superset relation.

\begin{table}[htbp]
\centering
\small
\setlength{\tabcolsep}{5pt}
\renewcommand{\arraystretch}{1.05}
\begin{tabularx}{\textwidth}{@{}l>{\raggedright\arraybackslash}X@{}}
\toprule
\textbf{Group} & \textbf{Checks} \\
\midrule
Structure and contracts & \gatename{basic\_structure}, \gatename{levels},
\gatename{level\_runtime}, \gatename{test\_contract},
\gatename{rubric\_contract}, \gatename{asset\_manifest},
\gatename{asset\_runtime\_references} \\
Build completion & \gatename{scenes\_implemented},
\gatename{assets\_generated}, \gatename{asset\_integration} \\
Execution evidence & \gatename{tests\_passing}, \gatename{test\_coverage},
\gatename{verify\_report}, \gatename{visual\_verified} \\
Alignment axes & \gatename{visual\_alignment}, \gatename{gameplay\_alignment},
\gatename{physics\_alignment}, \gatename{element\_alignment},
\gatename{layout\_alignment}, \gatename{visual\_diff},
\gatename{scene\_coverage} \\
Asset integrity & \gatename{bbox\_coverage}, \gatename{crop\_coverage},
\gatename{aspect\_drift}, \gatename{matting\_complete},
\gatename{no\_fx\_spritesheets}, \gatename{subject\_margin},
\gatename{slot\_validation} \\
Genre and interaction & \gatename{interaction\_contract},
\gatename{ui\_contract}, \gatename{ui\_layout},
\gatename{tower\_defence\_contract} \\
\bottomrule
\end{tabularx}
\caption{Checks in \gatename{iteration\_ready}, grouped by function.}
\label{tab:iterchecks}
\end{table}

\gatename{step\_done} is cumulative over steps 2 through 7. It always runs
\gatename{basic\_structure} and \gatename{pipeline\_state}; step~2 adds the five
scaffold-boundary checks, step~3 \gatename{level\_runtime},
\gatename{scenes\_implemented}, and \gatename{assets\_generated}, step~4
\gatename{asset\_integration}, \gatename{entity\_scale},
\gatename{ui\_proportion}, and \gatename{slot\_validation}, step~5 nothing
(the verification entry boundary), step~6 \gatename{visual\_verified}, and
step~7 \gatename{release\_waivers}, \gatename{playthrough\_recorded},
\gatename{verify\_report} in a full mode, and \gatename{performance}, for 19.

\subsection{The Gate Ladder}

\begin{figure}[!ht]
\centering
\includegraphics[width=0.94\linewidth]{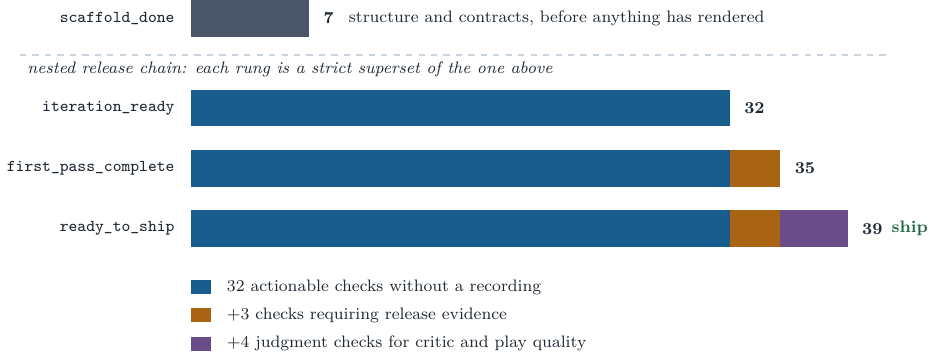}
\caption{Composite gate sizes, from the seven-check scaffold boundary to the
39-check release gate.}
\label{fig:ladder}
\end{figure}

\Cref{fig:ladder} draws the four composites to scale. They sit at distinct cost
boundaries: \gatename{scaffold\_done} after scaffolding,
\gatename{iteration\_ready} on every fix round, \gatename{first\_pass\_complete}
at the first full-mode pass, and \gatename{ready\_to\_ship} only on release.
Each is all-or-nothing --- no partial credit, no aggregate score to optimize ---
and when a repair or visual budget is exhausted the pipeline stops with the
unresolved list rather than continuing unbounded.

\FloatBarrier

\subsection{Threshold Reference}

All values below are the defaults in \artifact{gca.config.yaml}, the sole
authority: no agent specification, gate implementation, or reference document
restates them.

\begingroup
\small
\setlength{\tabcolsep}{5pt}
\renewcommand{\arraystretch}{1.40}
\begin{longtable}{@{}>{\raggedright\arraybackslash}p{6.3cm}>{\raggedright\arraybackslash}p{2.35cm}>{\raggedright\arraybackslash}p{7.15cm}@{}}
\caption{Default pipeline, gate, provider, and verification thresholds.}
\label{tab:thresholds}\\
\toprule
\textbf{Key} & \textbf{Value} & \textbf{Effect} \\
\midrule
\endfirsthead
\toprule
\textbf{Key} & \textbf{Value} & \textbf{Effect} \\
\midrule
\endhead
\multicolumn{3}{@{}l}{\emph{Iteration budget and evidence freshness} (\cfgkey{pipeline})} \\
\cfgkey{min\_iterations} / \cfgkey{max\_iterations} & 1 / 2 & Required fix rounds, and the hard cap \\
\cfgkey{max\_consecutive\_blocks} & 5 & Safety valve ahead of the budget \\
\cfgkey{min\_screenshots} / \cfgkey{min\_unique\_screenshots} & 3 / 3 & Distinct screenshots required as evidence \\
\cfgkey{staleness\_tolerance\_seconds} & 60 & Clock-skew slack on freshness \\
\cfgkey{max\_analysis\_age\_minutes} & 30 & Analyses expire after this \\
\cfgkey{min\_iteration\_gap\_seconds} & 120 & Minimum spacing between iterations \\
\cfgkey{max\_counter\_increment} & 1 & Guarded counters advance by at most one \\
\cfgkey{audit\_tolerance} & 1 & Slack between audit events and counters \\
\addlinespace[2pt]
\multicolumn{3}{@{}l}{\emph{Play, polish, and rubric floors} (\cfgkey{gates}, \cfgkey{gates.rubric})} \\
\cfgkey{play\_quality\_min\_accuracy} & 0.8 & Play-agent weighted accuracy floor \\
\cfgkey{polish.visual\_rounds\_min} & 1 & Visual evidence rounds at release \\
\cfgkey{polish.gameplay\_rounds\_min} & 1 & Gameplay evidence rounds at release \\
\cfgkey{polish.require\_stress\_tested} & false & Optional stress evidence \\
\cfgkey{intention\_min} & 0.9 & Does the video show the intended game \\
\cfgkey{visual\_min} / \cfgkey{interaction\_min} & 0.7 / 0.8 & Visual quality and interaction legibility \\
\addlinespace[2pt]
\multicolumn{3}{@{}l}{\emph{Performance} (\cfgkey{gates.performance})} \\
\cfgkey{min\_fps\_p50} / \cfgkey{min\_fps\_min} & 24 / 10 & Frame-rate floors \\
\cfgkey{max\_boot\_seconds} & 10 & Boot budget \\
\cfgkey{max\_long\_task\_count} & 20 & Long-task occurrences \\
\cfgkey{max\_long\_task\_total\_ms} & 2000 & Long-task total \\
\cfgkey{max\_memory\_growth\_mb} & 50 & Memory-leak detector \\
\cfgkey{max\_object\_growth\_count} & 250 & Entity-leak detector \\
\cfgkey{max\_object\_slope\_count\_per\_second} & 10 & Leak growth rate \\
\addlinespace[2pt]
\multicolumn{3}{@{}l}{\emph{Visual loop} (\cfgkey{gates.visual})} \\
\cfgkey{max\_rounds} & 2 & Visual budget, separate from the fix budget \\
\cfgkey{max\_attempts} / \cfgkey{unavailable\_policy} & 1 / warn & One request per comparison; outage warns \\
\cfgkey{method} & paired\_vlm & Per-element pairing strategy \\
\cfgkey{reference} & concept\_crop & Compared against the concept crop \\
\cfgkey{crop\_padding\_px} / \cfgkey{min\_element\_px} & 6 / 16 & Crop padding; smaller elements are skipped \\
\addlinespace[2pt]
\multicolumn{3}{@{}l}{\emph{Entity scale, framing, and UI proportion} (\cfgkey{gates.scale}, \cfgkey{gates.subject\_margin}, \cfgkey{gates.ui\_scale})} \\
\cfgkey{min\_entity\_px} / \cfgkey{max\_entity\_px} & 12 / 4096 & Rendered content height bounds \\
\cfgkey{body\_overflow\_tolerance} & 0.15 & Collision body may exceed art by 15\% \\
\cfgkey{body\_min\_ratio} & 0.30 & Collision body at least 30\% of art \\
\cfgkey{size\_class\_tolerance} & 0.10 & Slack on relative size ordering \\
\cfgkey{subject\_margin} top, bottom, left, right & 0.05 & Clear margin fraction per edge \\
\cfgkey{min\_px} / \cfgkey{max\_texture\_px} & 8 / 4096 & Rendered and source size bounds \\
\cfgkey{aspect\_tolerance} & 0.18 & Drift from the asset's own aspect ratio \\
\cfgkey{max\_canvas\_fraction} & 0.55\,/\,0.85\,/\,1.0 & Canvas share for \cfgkey{.hud}, \cfgkey{.panel}, \cfgkey{.banner} \\
\cfgkey{overlap\_min\_px} & 12 & Overlap on both axes counting as collision \\
\addlinespace[2pt]
\multicolumn{3}{@{}l}{\emph{Scene layout} (\cfgkey{gates.layout})} \\
\cfgkey{tolerance} & 0.04 & Position drift as a fraction of canvas diagonal \\
\cfgkey{min\_coverage} & 1.0 & All declared elements must render, by name \\
\cfgkey{overlap\_min\_px} / \cfgkey{overlap\_tolerance} & 12 / 0.12 & Rendered-bounds collision \\
\cfgkey{offscreen\_min\_px} & 8 & Permitted off-canvas overhang \\
\cfgkey{size\_drift\_tolerance} & 0.02 & Rendered size versus texture size \\
\addlinespace[2pt]
\multicolumn{3}{@{}l}{\emph{Interaction and anti-tunneling} (\cfgkey{gates.interaction}, \cfgkey{gates.tunnel})} \\
\cfgkey{min\_hit\_target\_px} & 24 & Measured hit body floor for pick targets \\
\cfgkey{tunnel.assumed\_fps} & 60 & Frame budget converting speed to travel \\
\cfgkey{tunnel.speed\_threshold} & 200 & Only faster bodies are checked \\
\cfgkey{tunnel.min\_body\_ratio} & 1.0 & Travel per frame stays under target body \\
\addlinespace[2pt]
\multicolumn{3}{@{}l}{\emph{Design scope caps} (\cfgkey{design})} \\
\cfgkey{max\_sprite\_entities} & 4 & Entities the design may plan \\
\cfgkey{max\_actions\_per\_entity} & 3 & Animated actions per entity \\
\cfgkey{max\_sprite\_action\_rows} & 15 & Total sprite-action rows \\
\cfgkey{max\_background\_layers} & 4 & Background layers \\
\cfgkey{max\_prop\_packs} / \cfgkey{max\_ui\_elements} & 3 / 4 & Prop packs and distinct UI element rows \\
\cfgkey{max\_concept\_images} & 6 & Concept mockups, hard cap 8 \\
\addlinespace[2pt]
\multicolumn{3}{@{}l}{\emph{Asset production and sanity} (\cfgkey{assets}, \cfgkey{gates.asset\_sanity})} \\
\cfgkey{image.max\_retries} & 3 & Closed-loop first-frame retries \\
\cfgkey{image.retry\_delay\_seconds} & 10 & Backoff between attempts \\
\cfgkey{image.safe\_bg\_color} & \#FF00FF & Chroma-key fill for mattable types \\
\cfgkey{seedance.max\_quality\_candidates} & 1 & One take per chain by default \\
\cfgkey{seedance.transient\_max\_attempts} & 3 & Retries on transient failure only \\
\cfgkey{matting.backend} & auto & Prefer remote matting, fall back on outage \\
\cfgkey{sprite.fps} / \cfgkey{sprite.max\_frames} & 12 / 24 & Spritesheet sampling \\
\cfgkey{sprite.max\_sheet\_width} & 4096 & Texture-size limit \\
\cfgkey{vlm.temperature} & 0.0 & Deterministic judgment \\
\cfgkey{vlm.slot\_attempts\_per\_model} & 1 & One slot request, no retry \\
\cfgkey{min\_png\_dimension\_px} & 32 & Anti-placeholder per-axis floor \\
\cfgkey{min\_unique\_files} / \cfgkey{min\_total\_bytes} & 3 / 50000 & Diversity and size floors \\
\addlinespace[2pt]
\multicolumn{3}{@{}l}{\emph{Verification runtime} (\cfgkey{verify})} \\
\cfgkey{play\_max\_wall\_seconds} & 240 & Play-agent wall budget \\
\cfgkey{record\_duration\_seconds} & 30 & Recording length \\
\cfgkey{max\_parallel\_producers} & 5 & Concurrent producers \\
\cfgkey{max\_parallel\_provider\_producers} & 2 & Concurrent provider-backed producers \\
\cfgkey{skip\_video} & false & Release includes recording unless waived \\
\bottomrule
\end{longtable}
\endgroup

\clearpage
\section{Judgment Rubrics}
\label{app:rubrics}

Every rubric below is evaluated by the pinned vision-language model at
temperature zero, and is condensed here from the validator adapters.
Decomposing a judgment into named binary rubrics rather than a quality score
follows the finding that instruction-guided multimodal judging is most usable
when each question is narrow and explicit~\citep{ku2024viescore}.

\subsection{Sprite First-Frame Rubrics}

These run under \toolname{validate\_image} before any animation is generated,
because a defect in the first frame propagates into every frame built from it.

\begin{table}[htbp]
\centering
\small
\setlength{\tabcolsep}{5pt}
\renewcommand{\arraystretch}{1.40}
\begin{tabularx}{\textwidth}{@{}ll>{\raggedright\arraybackslash}X@{}}
\toprule
\textbf{ID} & \textbf{Name} & \textbf{Condition for passing} \\
\midrule
I1 & Identity match & Species, gender, era, art style, and notable features
match the description exactly; no substitutions or embellishments. \\
I2 & Hand-prop attachment & Every prop shows a visibly gripping hand or
forearm; a prop merely near the body fails, however plausibly placed. \\
I3 & Limb integrity & All expected limbs present, torso-connected, at realistic
joint angles; none missing, extra, duplicated, or clipping. \\
I4 & Foot anchor & Lowest visible point in the bottom 30\% of the canvas; a
floating or vertically centered figure fails. \\
I5 & In-frame & Character and props unclipped, with at least 20\% margin above
the head and 15\% each side for later motion. \\
I6 & Background pure & Uniform solid chroma-key color; no gradient, shading,
noise, secondary object, or border. \\
I7 & Single character & Exactly one full-body character; no duplicates, pose
grids, UI overlays, captions, or frames. \\
I8 & View consistency & View angle matches the request: side, three-quarter,
front, or back. \\
\bottomrule
\end{tabularx}
\caption{Image rubrics I1--I8, applied to sprite first frames.}
\label{tab:irubrics}
\end{table}

Projectile bodies use a separate set P1--P6, since a fireball has no feet (I4),
hand (I2), or limbs (I3) to check. P1 identity, P5 background purity, and P6
single object mirror their character counterparts. P2 body integrity requires
one coherent object and forbids a baked motion trail, trails being runtime
particle effects rather than art. P3 centered with margin requires at least 8\%
clearance on every edge, unlike I4, where feet may rest on the bottom edge. P4
travel direction checks orientation against the described heading, passing as
not applicable for symmetric bodies.

\subsection{Sprite Animation Rubrics}

The animation rubrics run under \toolname{validate\_asset\_video} across two
inputs: the raw video, where the solid background still makes margins
measurable, and the matted frames that ship. A take ships only if it clears
both groups. Projectile animations use PR1--PR6, replacing the two anatomy
rubrics R4 and R5 with one body-integrity check.

\begin{table}[htbp]
\centering
\small
\setlength{\tabcolsep}{5pt}
\renewcommand{\arraystretch}{1.40}
\begin{tabularx}{\textwidth}{@{}ll>{\raggedright\arraybackslash}X@{}}
\toprule
\textbf{ID} & \textbf{Name} & \textbf{Condition for passing} \\
\midrule
\multicolumn{3}{@{}l}{\emph{Framing group, judged on the raw video}} \\
R1 & In-frame & Nothing is clipped in any frame, including peak motion: every
swing, sweep, dash, and trail stays fully inside the canvas. Feet on the bottom
edge pass. \\
R2 & Scale locked & Size stays within roughly 10--15\% of the first frame,
judged late-against-first so creep is caught. Extending a limb passes; scaling
the body does not. \\
\addlinespace[2pt]
\multicolumn{3}{@{}l}{\emph{Content group, judged on the matted output}} \\
R3 & Identity preservation & Same colors, patterns, face, and proportions in
every frame; no drift, swap, or morphing. \\
R4 & Hand-prop attachment & Held items stay anchored to a hand or the body;
nothing floats, detaches, or hovers without a visible grip. \\
R5 & Limb integrity & No limb disappears, duplicates, or bends impossibly; a
limb erased by the matte fails exactly as one the model never drew. \\
R6 & Loop coherence & Cyclic actions loop seamlessly, last frame matching the
first; one-shot actions such as death or hit-react pass as not applicable. \\
R7 & Action fidelity & The motion visibly performs the requested action; a
static pose, wrong action, wrong direction, or scale drift alone fails. \\
\bottomrule
\end{tabularx}
\caption{Video rubrics R1--R7.}
\label{tab:rrubrics}
\end{table}

\subsection{Release Review}

\toolname{analyze\_video} scores a sealed recording of gameplay against
configured floors: intention at 0.9, whether the video shows the game the
design document describes; visual at 0.7; and interaction at 0.8, whether
player action and game response are legible. Intention is the strictest floor
in the system: a recording that does not show the intended game is a
wrong-artifact problem, not a quality problem.

The critic then sees screenshots, that analysis, the play report, the
verification report, the design document, and the quality rubric; it does not
see source code, and its default position is that the build is not ready.

\begin{table}[htbp]
\centering
\small
\setlength{\tabcolsep}{5pt}
\renewcommand{\arraystretch}{1.40}
\begin{tabularx}{\textwidth}{@{}l>{\raggedright\arraybackslash}X@{}}
\toprule
\textbf{Dimension} & \textbf{Scale, from 0.1 through 0.3 and 0.5 to 0.8+} \\
\midrule
\cfgkey{visual\_quality} & Incoherent; rough or accidental; intentional but
uneven; profile-coherent, polished \\
\cfgkey{gameplay\_mechanics} & No gameplay evidence at all; partially working;
mostly working; all designed mechanics visible \\
\cfgkey{completeness} & Multiple major features missing; one or two missing;
all present but rough; complete and refined \\
\cfgkey{polish} & No transitions or feedback; basic transitions; has juice;
feels polished \\
\cfgkey{asset\_fitness} & Missing, broken, or illegible; inconsistent or poorly
integrated; runtime-fit but rough; cohesive and correctly integrated \\
\bottomrule
\end{tabularx}
\caption{Critic scoring dimensions and anchors.}
\label{tab:critic}
\end{table}

The verdict rule is deliberately asymmetric: a build passes only when all five
dimensions score 0.7 or above, is rejected outright when any falls below 0.5,
and is rejected by default in between. This hedges against the verbosity and
self-enhancement biases of model judges~\citep{zheng2023judging}. Three
overrides sit above the scores: a play agent's suspected bug caps gameplay
mechanics at 0.3; weighted accuracy below 0.5 is a strong rejection signal; and
any build-health failure---failing to boot, a fatal error, an empty
screenshot---rejects outright before the scores are read.

\clearpage
\section{Agent Interface Specification}
\label{app:interfaces}

This is the reproducibility contract: what someone needs in order to
re-implement any single agent and have it participate correctly. Phase order
belongs to the orchestration skill and phase-internal craft to each agent
specification, so neither is repeated here. Agents are dispatched with a flat
set of \texttt{key: value} fields; field order carries no meaning and every
path is absolute. Capability comes from the tool allowlist and the boundary
hook rather than from the prompt, so a field naming a path the agent does not
own is simply inert.

\newlength{\aifagent}\setlength{\aifagent}{2.5cm}
\newlength{\aifin}\setlength{\aifin}{7.5cm}
\newlength{\aifout}\setlength{\aifout}{5.885cm}
\newcommand{\aifk}[2]{\par\hangindent=1.1em\hangafter=1\relax\emph{#1}: #2}
\newcommand{\aifnote}[1]{%
  \multicolumn{2}{>{\raggedright\arraybackslash\footnotesize}%
    p{\dimexpr\aifin+\aifout+2\tabcolsep\relax}@{}}{#1}}

\subsection{Phase Interfaces}

\textbf{Startup routing} runs in the orchestrator with no subagent. For the
rest of the run it fixes the values of \cfgkey{genre}, \cfgkey{design\_axes},
\cfgkey{interaction\_profile}, \cfgkey{quality\_profile}, and
\cfgkey{level\_model}, and it sets \cfgkey{level\_design\_required} exactly
when the level model is authored. Fields that recur across interfaces are
defined once in \cref{tab:fields} and only named below.

\begingroup
\small
\setlength{\tabcolsep}{5pt}
\renewcommand{\arraystretch}{1.4}
\setlength{\parskip}{0pt}
\begin{longtable}{@{}p{\aifagent}%
  >{\raggedright\arraybackslash}p{\aifin}%
  >{\raggedright\arraybackslash}p{\aifout}@{}}
\caption{Phase-agent dispatch interfaces.}
\label{tab:phase-interfaces}\\
\toprule
\textbf{Agent} & \textbf{Inputs} & \textbf{Returns} \\
\midrule
\endfirsthead
\toprule
\textbf{Agent} & \textbf{Inputs} & \textbf{Returns} \\
\midrule
\endhead
\agentname{design}
& \aifk{Required}{\cfgkey{game\_idea}, \cfgkey{genre}, \cfgkey{design\_axes},
  \cfgkey{interaction\_profile}, \cfgkey{quality\_profile},
  \cfgkey{level\_model}, \cfgkey{game\_dir}, \cfgkey{prompt\_path}}
& \cfgkey{status}, \cfgkey{gdd\_path}, \cfgkey{sections},
  \cfgkey{scene\_count}, \cfgkey{asset\_count}, \cfgkey{concept\_mockups},
  \cfgkey{scene\_coverage}, \cfgkey{ref\_crops}, \cfgkey{warnings},
  \cfgkey{issues} \\
& \aifnote{\emph{Writes}: the design document and multimodal evidence under
  \artifact{design/}.} \\
\addlinespace[3pt]
\agentname{level-design}
& \aifk{Dispatch}{authored levels only}
  \aifk{Mode}{\cfgkey{create}, or \cfgkey{revise} with structured
  \cfgkey{issues}}
  \aifk{Required}{\cfgkey{gdd\_path}, \cfgkey{genre}, \cfgkey{level\_model},
  \cfgkey{game\_dir}}
& \cfgkey{stage\_count}, \cfgkey{level\_files}, \cfgkey{has\_boss},
  \cfgkey{extensions}, \cfgkey{design\_change\_requests} \\
\addlinespace[3pt]
\agentname{scaffold}
& \aifk{Mode}{\cfgkey{scaffold}, or \cfgkey{reproject} with non-empty
  \cfgkey{changed}}
  \aifk{Required}{\cfgkey{gdd\_path}, \cfgkey{game\_dir},
  \cfgkey{config\_path}, \cfgkey{quality\_profile}, \cfgkey{level\_model}}
& \cfgkey{contracts}, \cfgkey{files\_created}, \cfgkey{stale\_owned\_paths},
  \cfgkey{stale\_receipt\_id} \\
& \aifnote{Reproject deletion is receipt-mediated: scaffold seals its own
  stale list, and the orchestrator can only consume that receipt, never supply
  or widen the paths.} \\
\addlinespace[3pt]
\agentname{implement}
& \aifk{Dispatch}{one scene}
  \aifk{Required}{\cfgkey{scene\_name}, \cfgkey{gdd\_path}, \cfgkey{game\_dir},
  \cfgkey{genre}, \cfgkey{interaction\_profile}, \cfgkey{quality\_profile},
  \cfgkey{level\_model}, \cfgkey{url}}
  \aifk{Optional}{\cfgkey{visual\_evidence}, on visual fix rounds only}
& \cfgkey{scene}, \cfgkey{files\_changed}, \cfgkey{tests}, \cfgkey{issues},
  \cfgkey{design\_change\_requests} \\
& \aifnote{\cfgkey{url} carries server identity for continuity; it does not
  authorize browser execution, and implement authors tests without running
  them.} \\
\addlinespace[3pt]
\agentname{assets}
& \aifk{Dispatch}{one batch}
  \aifk{Mode}{\cfgkey{generate} or \cfgkey{register-existing}}
  \aifk{Required}{\cfgkey{gdd\_path}, \cfgkey{genre},
  \cfgkey{quality\_profile}, \cfgkey{level\_model}, \cfgkey{game\_dir}, and a
  \cfgkey{batch} scope}
& \cfgkey{assets\_generated}, \cfgkey{assets\_registered}, \cfgkey{sprites},
  \cfgkey{maps}, \cfgkey{props}, \cfgkey{ui}, \cfgkey{metadata\_sidecars} \\
& \aifnote{Workers write only owned files and sidecars, never the design
  document or the manifest.} \\
\addlinespace[3pt]
\agentname{verify}
& \aifk{Mode}{\cfgkey{iterate}, \cfgkey{full}, or \cfgkey{gate}}
  \aifk{Required}{\cfgkey{game\_dir}, \cfgkey{url}, \cfgkey{mode},
  \cfgkey{quality\_profile}, \cfgkey{level\_model}}
  \aifk{Conditional}{\cfgkey{gate} for gate mode, \cfgkey{step} for step
  diagnostics, \cfgkey{verify\_run\_id} for every verify and final gate}
& \cfgkey{verify\_run\_id}, producer \cfgkey{checkpoints}, \cfgkey{tests},
  evidence paths, \cfgkey{verify\_report}, \cfgkey{pre\_critic\_gate} \\
& \aifnote{Each canonical pass is exactly one \toolname{run\_verify\_dag}
  call. A \cfgkey{changed} list may drive a fast diagnostic but never
  substitutes for the unfiltered suite.} \\
\addlinespace[3pt]
\agentname{critic}
& \aifk{Required}{\cfgkey{game\_dir}, \cfgkey{screenshots},
  \cfgkey{video\_analysis}, \cfgkey{play\_report}, \cfgkey{verify\_report},
  \cfgkey{gdd\_path}, \cfgkey{rubric\_path}, \cfgkey{quality\_profile}}
& \cfgkey{verdict}, \cfgkey{top\_issue}, \cfgkey{issue\_category},
  \cfgkey{scores} \\
& \aifnote{Receives no \cfgkey{ref\_dir} and reads only the listed evidence.
  Its report must copy the verification run identity verbatim.} \\
\addlinespace[3pt]
\agentname{asset-integration}
& \aifk{Mode}{\cfgkey{integrate}}
  \aifk{Required}{\cfgkey{gdd\_path}, \cfgkey{game\_dir}, \cfgkey{genre},
  \cfgkey{interaction\_profile}, \cfgkey{quality\_profile},
  \cfgkey{level\_model}, \cfgkey{url}}
  \aifk{Optional}{a \cfgkey{scope} of \cfgkey{config}, \cfgkey{preloads},
  \cfgkey{animations}, \cfgkey{visual}, or \cfgkey{verify}}
& \cfgkey{assets\_wired}, \cfgkey{display\_params},
  \cfgkey{asset\_integration\_report} \\
& \aifnote{Dispatched only after both build branches join and the manifest is
  finalized; seals its receipt before any verification epoch opens.} \\
\bottomrule
\end{longtable}
\endgroup

\subsection{Shared Field Dictionary}

\begin{table}[H]
\centering
\small
\setlength{\tabcolsep}{5pt}
\renewcommand{\arraystretch}{1.14}
\begin{tabularx}{\textwidth}{@{}l>{\raggedright\arraybackslash}X@{}}
\toprule
\textbf{Field} & \textbf{Meaning} \\
\midrule
\cfgkey{game\_dir} & Absolute game-project root; MCP tools take the same path
under the same name \\
\cfgkey{gdd\_path} & Absolute path to the game design document \\
\cfgkey{genre} & Startup-selected genre; downstream agents never re-route it \\
\cfgkey{quality\_profile} & One of \cfgkey{art\_directed}, \cfgkey{responsive},
\cfgkey{pixel\_art}, \cfgkey{minimal} \\
\cfgkey{level\_model} & One of \cfgkey{authored}, \cfgkey{procedural},
\cfgkey{board}, \cfgkey{none} \\
\cfgkey{interaction\_profile} & Startup-selected physics and interaction
route \\
\cfgkey{url} & The exact local URL verified by the server launcher \\
\cfgkey{ref\_dir} & Absolute phase-specific reference directory; narrows the
context an agent reads \\
\cfgkey{config\_path} & Absolute path to \artifact{gca.config.yaml} \\
\cfgkey{level\_files} & Complete ordered array of absolute level paths;
required only for authored levels \\
\bottomrule
\end{tabularx}
\caption{Fields shared across dispatch interfaces.}
\label{tab:fields}
\end{table}

\subsection{Retry and Continuation}

Internal retries belong to the producer or the MCP server and use configured
bounds. The governing rule is that an unknown task state means wait, not retry:
a timeout does not prove termination, and treating it as terminal is how
duplicate work and split-brain state arise.

Recovery reopens the existing workspace in resume mode and reconciles durable
run, build, verification, and in-flight task identities before dispatch
continues; guarded state and audit updates recover valid pending transitions.
Completed work is reused only when its identity and freshness still match,
execution continues from the first incomplete or stale stage, and
browser-backed work additionally requires a freshly validated endpoint. No
partial write or artifact from an earlier run becomes current evidence.

\begin{table}[H]
\centering
\small
\setlength{\tabcolsep}{5pt}
\renewcommand{\arraystretch}{1.14}
\begin{tabularx}{\textwidth}{@{}l>{\raggedright\arraybackslash}X@{}}
\toprule
\textbf{Phase} & \textbf{Continuation rule} \\
\midrule
Design, level design, scaffold & No automatic retry. Design converts a bounded
concept-generation failure into its documented warning fallback: complete
layout and mechanics ship without the image. \\
Implement & No automatic scene retry; partial issues enter the fix loop, and
resume targets only incomplete scenes. \\
Assets & Internal retries are bounded by configuration; a terminal partial may
continue with only the missing keys. \\
Asset integration & One full dispatch first; scoped continuations are for
confirmed timeouts only and run sequentially in the order config, preloads,
animations, visual, verify. \\
Verify & Resume the same epoch from checkpoints after a confirmed
interruption. Never restart completed nodes, and never open a fresh epoch for
a transport failure. A new run starts only after the build changes or the
previous one finalizes. \\
Critic & No automatic retry; a failed judgment blocks assessment rather than
defaulting to either verdict. \\
\bottomrule
\end{tabularx}
\caption{Per-phase retry and continuation policy.}
\label{tab:retry}
\end{table}

\subsection{Return Envelope and Gate Interface}

Every return carries a \cfgkey{status}. \cfgkey{done} means validate the
outputs, persist completion, and continue; \cfgkey{partial} means retain the
valid outputs, record every issue, and continue only when safe;
\cfgkey{blocked} means report concrete issues and wait. Warnings are
non-blocking, and issues describe incomplete or blocked work. The orchestrator
never force-continues past a blocked return, never rewrites an agent's return,
and never treats an in-flight timeout as terminal. Each prohibition closes a
route by which an orchestrator pressed to make progress could manufacture the
appearance of completion, the exact failure the evidence architecture exists to
prevent.

The orchestrator may call \toolname{gate\_check} directly, and
\agentname{verify} in gate mode uses the same interface. Every response returns
\cfgkey{blocked} plus a structured \cfgkey{failures[]} in which each record
carries the failing \cfgkey{check}, routing hints naming the \cfgkey{axis} and
owning \cfgkey{agent}, and a message; grouping by \cfgkey{agent} yields the
parallel fixer dispatch described in \cref{sec:system}. Each
\gatename{step\_done} call writes its receipt to a step-numbered path, so one
phase's diagnostic cannot overwrite another's. Three advisory diagnostics
(\toolname{check\_ui\_contract}, \toolname{check\_td\_contract},
\toolname{check\_interaction\_contract}) expose deterministic contract gates
for fast mid-build feedback without opening an epoch. They are explicitly not
release authorities: the shipping decision is \gatename{ready\_to\_ship} with a
matching verification run identity, and the stop hook trusts only that gate's
sealed result.

\clearpage
\section{Human Study Questionnaire}
\label{app:human-questionnaire}

\begingroup
\small
\setlength{\parskip}{0.30em}

\textbf{Participant instructions.}
Read the assigned prompt, then play Game A and Game B in the recorded order.
System identities are hidden behind randomized labels. Rate each game from
your own interaction and observation, without inferring quality from loading
time outside the game, file names, or perceived authorship. Complete one copy
for each of the 20 pairs in \cref{tab:eval-corpus}.

\textbf{Participant ID:} \rule{2.4cm}{0.4pt}\qquad
\textbf{Session ID:} \rule{2.2cm}{0.4pt}\qquad
\textbf{Date:} \rule{2.2cm}{0.4pt}

\textbf{Report task no.:} \rule{1.8cm}{0.4pt}\qquad
\textbf{Task/game name:} \rule{4.1cm}{0.4pt}

\textbf{Pair position:} \rule{0.8cm}{0.4pt}\,/\,20\qquad
\textbf{Randomization:} \rule{1.8cm}{0.4pt}\qquad
\textbf{Order:}\quad
\(\square\) A--B\quad
\(\square\) B--A

\textbf{Shared response scale.}\quad
\textbf{1} = clearly not met;\quad
\textbf{2} = mostly not met;\quad
\textbf{3} = partly met;\quad
\textbf{4} = mostly met;\quad
\textbf{5} = fully met.

\vspace{2pt}
\setlength{\tabcolsep}{3.5pt}
\renewcommand{\arraystretch}{1.14}
\begin{tabularx}{\linewidth}{@{}
  >{\raggedright\arraybackslash}p{2.65cm}
  >{\raggedright\arraybackslash}X
  >{\centering\arraybackslash}p{1.55cm}
  >{\centering\arraybackslash}p{1.55cm}@{}}
\toprule
\textbf{Rating} & \textbf{Participant question} &
\makecell{\textbf{Game A}\\\textbf{(1--5)}} &
\makecell{\textbf{Game B}\\\textbf{(1--5)}} \\
\midrule
Requirement Satisfaction &
How fully does the game realize the mechanics, content, progression, visible
feedback, and outcomes required by the prompt? &
\rule{1.0cm}{0.4pt} & \rule{1.0cm}{0.4pt} \\

Core Mechanics &
Do the controls, rules, state transitions, success/failure conditions, and
restart behavior work during play? &
\rule{1.0cm}{0.4pt} & \rule{1.0cm}{0.4pt} \\

Content Depth &
Are the requested modes, levels, progression, challenge variation, and
terminal content present? &
\rule{1.0cm}{0.4pt} & \rule{1.0cm}{0.4pt} \\

Functional Visuals &
Can you see the important game state, HUD values, action feedback, and
settlement information when needed? &
\rule{1.0cm}{0.4pt} & \rule{1.0cm}{0.4pt} \\

Art \& Presentation &
Beyond basic function, how coherent, readable, well composed, and visibly
finished is the presentation? &
\rule{1.0cm}{0.4pt} & \rule{1.0cm}{0.4pt} \\
\bottomrule
\end{tabularx}

\vspace{5pt}
\textbf{Overall preference.}\quad
Considering prompt fulfillment, play, and presentation together:

\(\square\) Game A\qquad
\(\square\) Game B\qquad
\(\square\) No preference

\vspace{5pt}
\textbf{Issue/comment.}\quad
Note any blocked action or concrete interaction/visual observation that
influenced your ratings or preference.

\vspace{5pt}
\rule{\linewidth}{0.4pt}

\vspace{10pt}
\rule{\linewidth}{0.4pt}

\vspace{5pt}
\footnotesize
Game A and Game B ratings are independent observations. Preference is
categorical and is analyzed separately; it does not replace either five-item
rating vector.
\endgroup

\end{document}